\documentclass{article}

\usepackage{microtype}
\usepackage{graphicx}
\usepackage{subfigure}
\usepackage{booktabs} % for professional tables
\usepackage{svg}

\usepackage{hyperref}
\usepackage{algpseudocode}
\usepackage{url}
\usepackage{subcaption}
\usepackage{caption}
\usepackage{wrapfig}
\usepackage{listings}
\usepackage{algpseudocode}
\newcommand{\greencheck}{{\color{green}\checkmark}} 
\newcommand{\redcross}{{\color{red}$\times$}}

\expandafter\def\expandafter\normalsize\expandafter{%
    \normalsize%
    \setlength\abovedisplayskip{4pt}%
    \setlength\belowdisplayskip{4pt}%
    \setlength\abovedisplayshortskip{-4pt}%
    \setlength\belowdisplayshortskip{4pt}%
}

\usepackage[accepted]{icml2024}

\usepackage{amsmath}
\usepackage{amssymb}
\usepackage{mathtools}
\usepackage{amsthm}

\usepackage[capitalize,noabbrev]{cleveref}

\theoremstyle{plain}

\theoremstyle{definition}

\theoremstyle{remark}

\usepackage[textsize=tiny]{todonotes}

\icmltitlerunning{Language Agent Tree Search Unifies Reasoning, Acting, and Planning in Language Models}

\begin{document}

\twocolumn[
\icmltitle{Language Agent Tree Search Unifies Reasoning, Acting, and Planning in Language Models}
%~\yxw{I think unify is better than improve - unify suggests more important contrition}\kai{modified}



\icmlsetsymbol{equal}{*}

\begin{icmlauthorlist}
\icmlauthor{Andy Zhou}{uiuc,lapis}
\icmlauthor{Kai Yan}{uiuc}
\icmlauthor{Michal Shlapentokh-Rothman}{uiuc}
\icmlauthor{Haohan Wang}{uiuc}
\icmlauthor{Yu-Xiong Wang}{uiuc}
\end{icmlauthorlist}

\icmlaffiliation{uiuc}{University of Illinois Urbana-Champaign.}
\icmlaffiliation{lapis}{Lapis Labs}

\icmlcorrespondingauthor{Andy Zhou}{andyz3@illinois.edu}

\icmlkeywords{Machine Learning, Large Language Models, LM Agents, LM Planning}

\vskip 0.3in
]



\printAffiliationsAndNotice{} % otherwise use the standard text.
\begin{abstract}
While language models (LMs) have shown potential across a range of decision-making tasks, their reliance on simple acting processes limits their broad deployment as autonomous agents. In this paper, we introduce Language Agent Tree Search (LATS) -- \emph{the first general} framework that \emph{synergizes} the capabilities of LMs in reasoning, acting, and planning. By leveraging the in-context learning ability of LMs, we integrate Monte Carlo Tree Search into LATS to enable LMs as agents, along with LM-powered value functions and self-reflections for proficient exploration and enhanced decision-making. A key feature of our approach is the incorporation of an environment for external feedback, which offers a more deliberate and adaptive problem-solving mechanism that surpasses the constraints of existing techniques. Our experimental evaluation across diverse domains, including programming, interactive question-answering (QA), web navigation, and math, validates the effectiveness and generality of LATS in decision-making while maintaining competitive or improved reasoning performance. Notably, LATS achieves state-of-the-art pass@1 accuracy (92.7\%) for programming on HumanEval with GPT-4 and demonstrates gradient-free performance (average score of 75.9) comparable to gradient-based fine-tuning for web navigation on WebShop with GPT-3.5. Code can be found at \url{https://github.com/lapisrocks/LanguageAgentTreeSearch}.
\end{abstract}

\section{Introduction}

General autonomous agents capable of reasoning and decision-making in a variety of environments \citep{wooldridge1995intelligent} have been of longstanding interest in the field of artificial intelligence. While this has traditionally been studied in reinforcement learning, the recent rise of language models (LMs) \citep{GPT3, chowdhery2022palm, touvron2023llama, openai2023gpt4} with strong reasoning and general adaptability offers an alternative paradigm. Not only have LMs excelled in standard natural language processing (NLP) tasks such as summarization \citep{nallapati2016abstractive} and language inference \citep{bowman2015large}, but they have also been adapted to an increasingly diverse set of tasks that often require advanced common-sense reasoning or quantitative skills \citep{cobbe2021training, saparov2022language}. In addition, LMs are capable of performing in complex environments that involve knowledge and reasoning, such as web navigation \citep{yao2022webshop, deng2023mind2web}, tool-use \citep{schick2023toolformer}, and open-ended games \citep{fan2022minedojo}. 

% \begin{wrapfigure}{r}{0.5\textwidth}
%   \centering
%   \vspace{-0.2in}
%   \includegraphics[width=0.4\textwidth]{figures/teaser2.pdf}
%   \caption{An overview of LATS. LATS uses an external environment and self-reflection to improve	reasoning and decision-making.}
%   \vspace{-0.2in}
% \end{wrapfigure}
\begin{figure}
    \centering
    % \includesvg[width=0.36\textwidth]{figures/lats_teaser.svg}
    \includegraphics[width=0.38\textwidth]{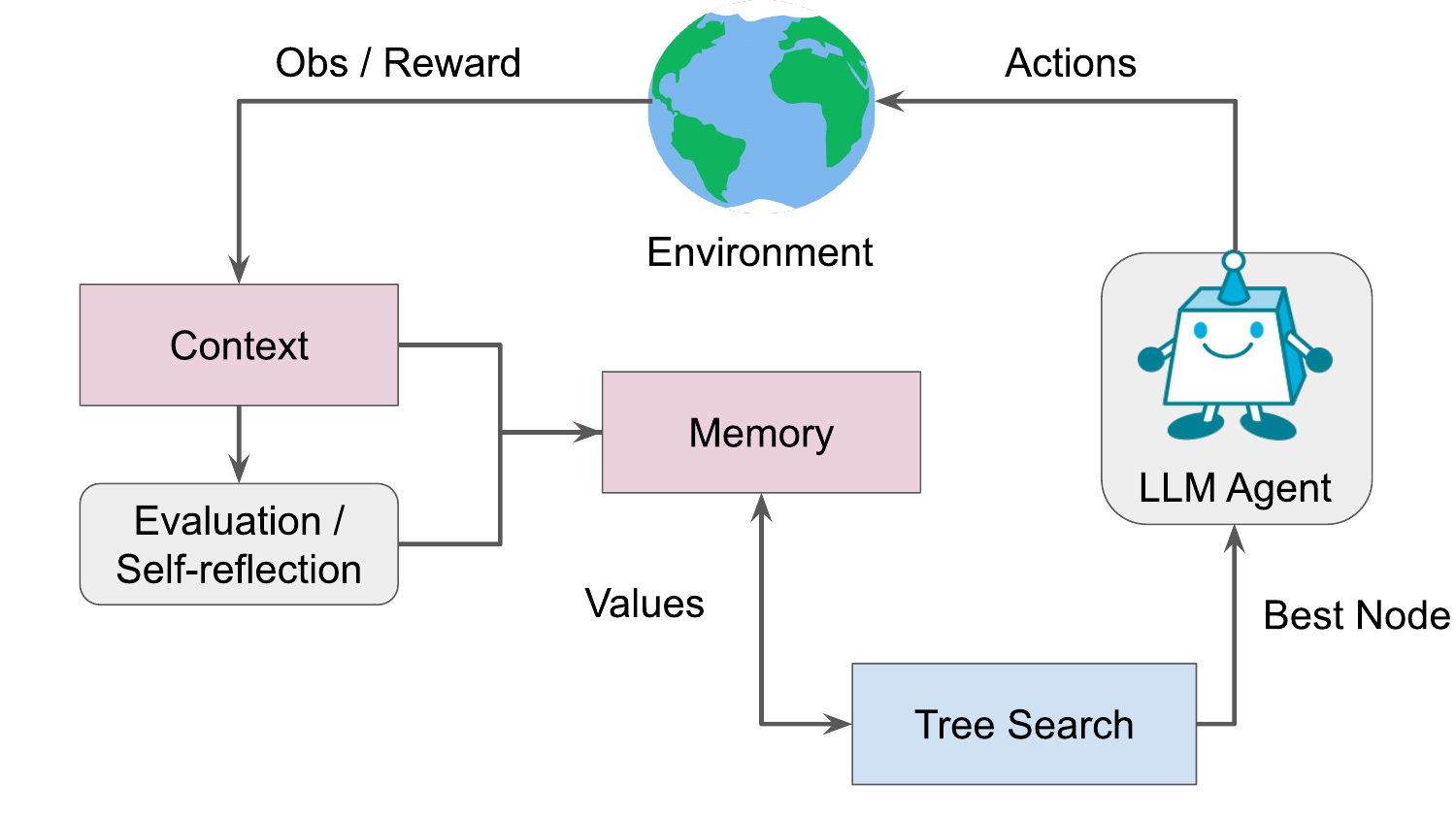}
    \vspace{-0.1in}
    \caption{Overview of LATS. Serving as a unified framework, LATS leverages an external environment and an MCTS-based search algorithm to improve reasoning and decision-making.}
    \label{fig:teaser}
    \vspace{-0.2in}
\end{figure}
%~\yxw{Fig. 1 is not referred in the main text.}\az{added in paragrph4}

Reasoning and acting abilities have been further improved by prompting techniques that augment LMs with feedback or observations from an external environment, as exemplified by ReAct \citep{yao2023react} and other work \citep{gao2022pal, shinn2023reflexion}. This eliminates the need to rely entirely on the base abilities of LMs, enhancing them through external tools or semantic feedback. Despite such strengths, these methods are reflexive and fall short of humans' deliberate and thoughtful decision-making characteristics to solve problems \citep{Sloman1996TheEC,evans2010intuition}. In particular, they fail to consider multiple reasoning paths or to plan ahead. Recent search-guided LM work \citep{xie2023decomposition, yao2023tree, hao2023reasoning} addresses this issue by searching over multiple reasoning chains. While enabling planning, such methods operate in isolation, lacking the incorporation of external feedback that can improve reasoning. 
\vspace{0.04in}

To overcome these challenges, we propose Language Agent Tree Search (LATS) -- a \emph{unified} framework for decision-making and reasoning with language models. As illustrated in Fig.~\ref{fig:teaser}, LATS \emph{synergizes LM reasoning, acting, and planning} strategies by expanding ReAct \citep{yao2023react} into a search over a combinatorial space of possible reasoning and acting steps. This effort is nontrivial -- adapting search algorithms to language agents and shifting from non-interactive tasks to interactive ones requires a substantial novel design on nodes, prompts, and search algorithms. In particular, nodes and prompts must effectively store and retrieve external feedback, with the search algorithm incorporating this information into useful heuristics for value assignment. Indeed, our empirical evaluation, as demonstrated on HotPotQA \citep{yang2018hotpotqa} in Sec.~\ref{sec:HotPotQA}, reveals that a simple combination of existing methods is inadequate, even failing to surpass internal reasoning performance, despite having access to the ground truth answer from the environment.
\vspace{0.04in}

Our \emph{key insight} underpinning LATS is adapting Monte Carlo Tree Search (MCTS), inspired by its success in model-based reinforcement learning \citep{silver2017mastering} and \textit{the observation that many LM tasks allow reverting to earlier steps}, to language agents, repurposing pretrained LMs as agents with LM-powered value functions and self-reflections for cleverer exploration. Leveraging the general capabilities and in-context learning abilities of modern LMs, we use language as an interface between each component, allowing LATS to adapt planning to environmental conditions \emph{without additional training}. To the best of our knowledge, LATS is \emph{the first framework} that incorporates reasoning, acting, and planning to enhance LM performance. Notably, LATS doubles the performance of ReAct \citep{yao2023react} on HotPotQA \citep{yang2018hotpotqa} and raises the average score by $22.1$ on WebShop \citep{yao2022webshop} with GPT-3.5. When used with GPT-4, LATS achieves a $92.7$ Pass@1 rate on HumanEval \citep{chen2021evaluating}, setting the state of the art. 

\vspace{0.04in}

Our \textbf{contributions} are the following: 1) We introduce LATS, a framework based on Monte Carlo Tree Search to construct the best trajectory from sampled actions, enabling more flexible and adaptive problem-solving compared with reflexive prompting methods. 2) We propose a novel value function that guides the search process and incorporates successful heuristics such as self-refinement and self-consistency. 3) By integrating external feedback and self-reflection, LATS enhances model sensibility and enables agents to learn from experience, surpassing reasoning-based search methods. Through experiments across diverse domains, including programming, interactive question-answering (QA), web navigation, and math, we demonstrate the versatility of LATS for enhancing autonomous reasoning and decision-making.
%To summarize, 
\section{Related Work}
\vspace{-0.06in}
\begin{table*}
    \centering
    \begin{tabular}{lccccc}
    \toprule
    \textbf{Approach} & 
    \textbf{Reasoning} &
    \textbf{Acting} &
    \textbf{Planning} &
    \textbf{Self-} &
    \textbf{External} \\
    & & & & \textbf{Reflection} & \textbf{Memory} \\
    \midrule
    CoT \citep{wei2022chain} & \greencheck & \redcross & \redcross & \redcross & \redcross \\
    ReAct \citep{yao2023react} & \greencheck & \greencheck & \redcross & \redcross & \redcross \\
    ToT \citep{yao2023tree} & \greencheck & \redcross & \greencheck & \greencheck & \greencheck \\
    RAP \citep{hao2023reasoning} & \greencheck & \redcross & \greencheck & \redcross & \greencheck \\
    Self-Refine \citep{madaan2023selfrefine} & \greencheck & \redcross & \redcross & \greencheck & \redcross \\
    Beam Search \citep{xie2023decomposition} & \greencheck & \redcross & \redcross & \greencheck & \redcross \\
    Reflexion \citep{shinn2023reflexion} & \greencheck & \greencheck & \redcross & \greencheck & \greencheck \\
    \textbf{LATS (Ours)} & \greencheck & \greencheck & \greencheck & \greencheck & \greencheck \\
    \bottomrule
    \end{tabular}

    \caption{Summary of related work on reasoning, acting, and planning. LATS is the \emph{first} work incorporating designs from \emph{all} three domains, allowing broad applicability in all corresponding tasks. We refer to reasoning as LM internal reasoning, acting as external decision-making, planning as the use of a search algorithm, self-reflection as the use of LM-generated feedback, and external memory as storing past text context for future updates of the solution.}
    \vspace{-0.1in}
    \label{table:related_work}
\end{table*}

\textbf{LMs for reasoning.} For LMs, reasoning involves decomposing complex inputs into sequential intermediate steps towards a final answer \citep{cobbe2021training}, demonstrated with chain-of-thought (CoT) prompting \citep{wei2022chain} and its variants~\citep{wei2022chain, kojima2022large, wang2022self}. However, these methods, which create chains autoregressively in a single step, often suffer from error propagation as the number of steps increases \citep{guo2018long,chen2022program}, due to compound errors. Various advancements aim to mitigate this issue; some approaches, such as self-consistency~\citep{wang2022self}, employ majority voting over sampled chains, while others focus on multi-step decomposition, such as least-to-most prompting~\citep{zhou2022least}. Recently, CoT has been improved with search algorithms \citep{yao2023tree,hao2023reasoning,besta2023graph} that can sample trajectories more effectively. Tree-of-thought (ToT) prompting \citep{yao2023tree} uses DFS or BFS-based (depth/breadth-first) search guided by an LM-generated heuristic, while reasoning via planning (RAP) \citep{hao2023reasoning} uses MCTS with rollouts simulated by LMs. However, they rely solely on LM internal knowledge and cannot adapt to useful external feedback.
%typically

\textbf{LMs for acting.} The strong reasoning and common-sense abilities of LMs have been further adapted for decision-making or acting tasks as a policy model in interactive environments. In robotics, LMs have been employed as high-level controllers of control policies \citep{ahn2022do, huang2022inner, driess2023palme}. Similar work \citep{openai2022vpt, wang2023voyager} has also adapted LM agents to complex multimodal games such as Minecraft \citep{guss2019minerl, fan2022minedojo}. LMs are particularly useful in text-based environments \citep{liu2018reinforcement,shridhar2020alfworld,liu2023agentbench}, where acting-based prompting techniques such as ReAct \citep{yao2023react} have seen success. Similar to CoT, ReAct is limited by its simplicity and cannot effectively adapt to environment conditions. Many extensions have been proposed to address this issue, including self-refine \citep{madaan2023selfrefine} and Reflexion \citep{shinn2023reflexion}, which use self-improvement to enhance reasoning and decision-making, and AdaPlanner \citep{sun2023adaplanner}, which incorporates both positive and negative feedback. However, these methods focus on refining an individual trajectory and do not consider alternative choices at each step. In addition, recent work \citep{huang2023large} has suggested that LMs cannot self-correct their internal reasoning, making it critical to use external feedback. Alternatively, to pure decision-making environments, the reasoning and practical abilities of LMs have been enhanced by providing access to external tools, such as APIs, search engines, calculators, and other models \citep{schick2023toolformer, shen2023hugginggpt, surís2023vipergpt}. We summarize prior work in Tab.~\ref{table:related_work}.

\textbf{Tree-based search.} Tree-based search, where multiple branches of outcomes are explored during search, is widely used in many planning algorithms~\citep{swiechowski2023monte, lavalle2001rapidly} and reinforcement learning (RL)~\citep{hafner2019learning, du2023learning, wu2023daydreamer} algorithms for its good exploration-exploitation trade-off. Note that though tree-based search necessitates an environment model that can expand from an arbitrary state~\citep{vodopivec2017monte}, often requiring extra training in RL~\citep{hafner2023mastering}, such a problem \emph{does not} exist for most LM tasks. This is because we can conveniently revert to any state by setting the input to be the context and the corresponding previous output from the LM for many tasks. Thus, we operate on the tree-based framework and use MCTS~\citep{swiechowski2023monte} to fully unlock the potential of LMs. In addition, we avoid the cost of training a value function over language descriptions by leveraging the in-context learning \citep{GPT3} abilities of LMs. Concurrent work \cite{liu2023reason} also explores combining search algorithms with LM agents but uses an off-the-shelf search algorithm, which may not be optimal for LMs. Finally, following \citet{yao2023tree} and \citet{hao2023reasoning}, we note that we use \textit{planning} and \textit{search algorithms} interchangeably in this paper.%{\color{red}Concurrent work \cite{liu2023reason} also explores combining search algorithms with LM agents but uses an off-the-shelf search algorithm. Finally, we note that we do not distinguish between \textit{planning} and the use of a \textit{search algorithm} in this paper, following \citet{yao2023tree} and \citet{hao2023reasoning}.}
%~\yxw{consider changing the title to include LMs for planning?} \kai{But this paragraph is about tree-based search, e.g., MCTS in general and does not focus on NLP task}~\yxw{The red text here needs revising, hard to parse.}\kai{revised}

\section{Preliminaries}

\subsection{Problem Setting and Prompting}

We first define our problem and outline a few established methods that leverage language models for reasoning \emph{or} decision-making. In LM reasoning or decision making, we are given an input $x$ in natural language and a pretrained language model $p_\theta(x)$ parameterized by $\theta$; our goal is to generate a final output $y\sim p_\theta(x)$ that corresponds to the answer (reasoning) or completes the task (decision-making). Both $x$ and $y$ are language \textit{sequences}, which are comprised of a list of \textit{tokens} (the basic elements of natural language, often words), denoted as \( x = (x[1], \dots, x[l_x]) \) and \( y = (y[1], \dots, y[l_y]) \) where $l_x$ and $l_y$ are the length. The LM decodes text autoregressively, i.e., without other inputs, the probability for an LM to generate a sequence $y$ is given by \(p_\theta(x) = \prod_{i=1}^{l_x} p_\theta(x[i] | x[1 \dots i-1]) \). Usually, to improve reasoning, \textit{prompts} are provided along with the input $x$, which are specific instructions or few-shot input-output examples. We denote the generic process where an input \( \texttt{prompt}_{\mathrm{IO}}(x) \) is transformed into an output \( y \) by LM: \( y \sim p_\theta(\texttt{prompt}_{\mathrm{IO}}(x)) \).
%, where \( \texttt{prompt}_{IO}(x) \) denotes the input \( x \).
%large ~\yxw{this is not consistent with the following usage, should it be $y\sim p_\theta(y|x)$?}\kai{modified} ~\yxw{what is n? - not defined; also do x and y have to be of the same length?}\kai{Andy - please address}\az{done} ~\yxw{why LLM generates x? x is input.}\az{mistake, changed to y} ~\yxw{what does it mean by `improve LM'?}\kai{explained}
%Before describing LATS, w

\textbf{Chain-of-thought (CoT) prompting}~\citep{wei2022chain} caters to scenarios where the direct mapping from \( x \) to \( y \) is intricate, e.g., when \( x \) is from a mathematical query or challenging question. It hinges on creating \textit{thoughts} \( z_1, \dots, z_l \) that act as stepping stones between \( x \) and \( y \); each thought $z_i$ is a language sequence. To employ CoT prompting, thoughts are extracted sequentially as \( z_i \sim p_\theta^{\mathrm{CoT}}(x, z_{1\cdots i-1}) \), with the final output being \( y \sim p_\theta^{\mathrm{CoT}}(x, z_{1 \cdots l}) \).
%~\yxw{$p_\theta^\mathrm{CoT}$ - use mathrm for text-based sub-script/superscript}\kai{Addressed} is introduced to such as This method

\textbf{Tree-of-thought (ToT) prompting}~\citep{yao2023tree} extends CoT prompting by exploring multiple reasoning paths over thoughts. It frames problems as a search over a tree, where each node $s=[x, z_{1\cdot i}]$ represents a partial solution state comprising the original input $x$ and the thought sequence $z_{1\cdots i}$. Thoughts $z_{i}$ are generated by proposal or sampling with CoT \( z_i \sim p_\theta^\mathrm{CoT}(x, z_{1\cdots i-1}) \). Search algorithms like depth-first (DFS) or breadth-first (BFS) search are used to systematically explore the tree, guided by heuristics based on LM evaluations $V(s)$ of each state.
%{ \yxw{I think this sentence can be made stronger - need to first mention this assumption is widely existing in ...; and then say we follow.}\az{moved to react section}

\textbf{ReAct}~\citep{yao2023react} extends language models to tasks where the mapping from \( x \) to \( y \) is enhanced by or requires interactions with an external environment, such as a game or API. This technique constructs an action space $\hat{A} = A \cup Z$ that adds permissible actions $a \in A$ to the reasoning traces $z \in Z$ from CoT. Observations $o$ from the environment are used to improve both reasoning and acting. To solve problems with ReAct, after each observation, actions are generated from $p_\theta$ sequentially as \( a_i \sim p_\theta^\mathrm{ReAct}(x, o_{1\cdots i-1}, a_{1\cdots i-1}) \), with the final output being \( y \sim p_\theta^\mathrm{ReAct}(x, o_{1\cdots l}, a_{1\cdots l}) \). In this paper, consistent with other LM agent methods such as ReAct and Reflexion \cite{shinn2023reflexion}, we focus on decision-making tasks \emph{where reverting between iterations is feasible}.

While the previously described prompting techniques improve LM performance on reasoning tasks, they falter on difficult tasks that involve multifaceted decision-making due to several shortcomings: 1) \textit{Flexibility}: Base prompting designs (CoT or ReAct) autoregressively sample from the LM, neglecting potential alternative continuations from specific states. 2) \textit{Sensibility}: Reasoning-based methods (CoT, RAP \cite{hao2023reasoning}, or ToT) rely solely on the internal representations of the LM and cannot consider external observations. This dependency risks fact hallucination and error propagation while setting a performance ceiling. 3) \textit{Adaptability}: Current planning strategies (RAP or ToT) use simple search algorithms such as BFS or cannot leverage environmental feedback to improve planning. Additionally, the agent is static and cannot reuse previous experience or learn from trial and error. While RAP also adopts MCTS, it is constrained to tasks where the LM can become a world model and accurately predict states. These shortcomings limit the ability of LMs to be deployed as general problem-solving agents and form the motivation for LATS.

\subsection{Monte Carlo Tree Search (MCTS)}
Monte Carlo Tree Search (MCTS) is a heuristic search algorithm that is proved successful on many decision-making environments, such as Atari~\citep{ye2021mastering} and Go~\citep{silver2016mastering}. MCTS builds a decision tree where every node in the tree is a state and edge is an action. MCTS runs for $k$ episodes; for each episode, it starts from the root (i.e., initial state) and iteratively conducts two steps to expand the tree: 1) \textit{Expansion}, where multiple children states $s$ are explored from the current parent state $p$ by sampling $n$ actions, and 2) \textit{Selection}, where the children with the highest UCT \textit{(Upper Confidence bounds applied to Trees)}~\citep{kocsis2006bandit} value is selected for expansion by the next iteration. The UCT of a child state $s$ is calculated as follows:
\begin{equation}
\small
UCT(s)=V(s)+w\sqrt{\frac{\ln N(p)}{N(s)}},
\label{eq:uct}
\end{equation}
where $N(s)$ is the number of visits to a node $s$, $V(s)$ is the value function (expected return) from the subtree of $s$, $w$ is the exploration weight, and $p$ is the parent node of $s$. When the end of an episode is reached, a \textit{backpropagation} is carried out: the return $r$ is used for updating every $V(s)$ along the path 
% see Sec.~\ref{sec:LATS} for formulas). 
with the formula $V(s)=\frac{V_{\text{old}}(s)(N(s)-1)+r}{N(s)}$, where $V_{\text{old}}(s)$ is the old value function. Normally, the major shortcoming of MCTS is that it requires an environment model to undo previous steps and form a searching tree, which could be a strong assumption. However, this limitation \emph{does not} exist for many LM tasks, as we can conveniently reset to any step by simply copy-pasting historical text input. Such a special property is the key motivation of our work. 
%~\yxw{is this sentence a bit redundant with the sentence above eq. 1?}~\kai{Removed}.often
% ------------------------------------------------

\section{Unifying Reasoning, Acting, and Planning}
\subsection{LM Agent} 
%~\yxw{why not follow the same order as the title?}\kai{Addressed}

Depending on the base prompting framework design, LATS supports sequential reasoning or decision-making tasks. At time step $t$, an agent receives an observation $o_t \in O$ from the environment and takes an action $a_t \in A$ following some policy $\pi(a_t | x, o_{1\cdots t-1}, a_{1\cdots t-1})$. We initialize the agent with $p_\theta$ to leverage the useful language representations of an LM as a base decision-maker. We follow the ReAct instantiation, in which the action space $\hat{A} = A \cup Z$ consists of both the space of permissible actions $A$ and the language space of reasoning traces $Z$. Actions directly affect the environment and result in observation, while thoughts are used to formalize decisions by organizing information, planning future actions, or injecting internal knowledge. The exact instantiation of the action space depends on the particular environment -- for decision-making tasks actions might consist of commands on a website, while for reasoning tasks the action space might be limited to a few external tools or APIs. In environments without feedback, such as reasoning tasks, we use CoT as the base prompting framework.
%~\yxw{would there be any confusion between t and i?}\kai{modified}

Instead of greedily decoding one trajectory or solution, we sample $n$ actions from $p_\theta$ using the current state. This is based on the intuition that for complex decision-making tasks, there is likely to be a range of potential trajectories or reasoning paths that are correct \citep{evans2010intuition}. Sampling a diverse set of candidates at each step mitigates the stochastic nature of LM text generation and enables greater exploration in both the decision-making and reasoning space. We wrap $p_\theta$ within our proposed search algorithm to deliberately construct the best trajectory from sampled actions.

\subsection{LATS}
\label{sec:LATS}
\begin{figure*}[t]
    \centering
    \includegraphics[width=\linewidth]{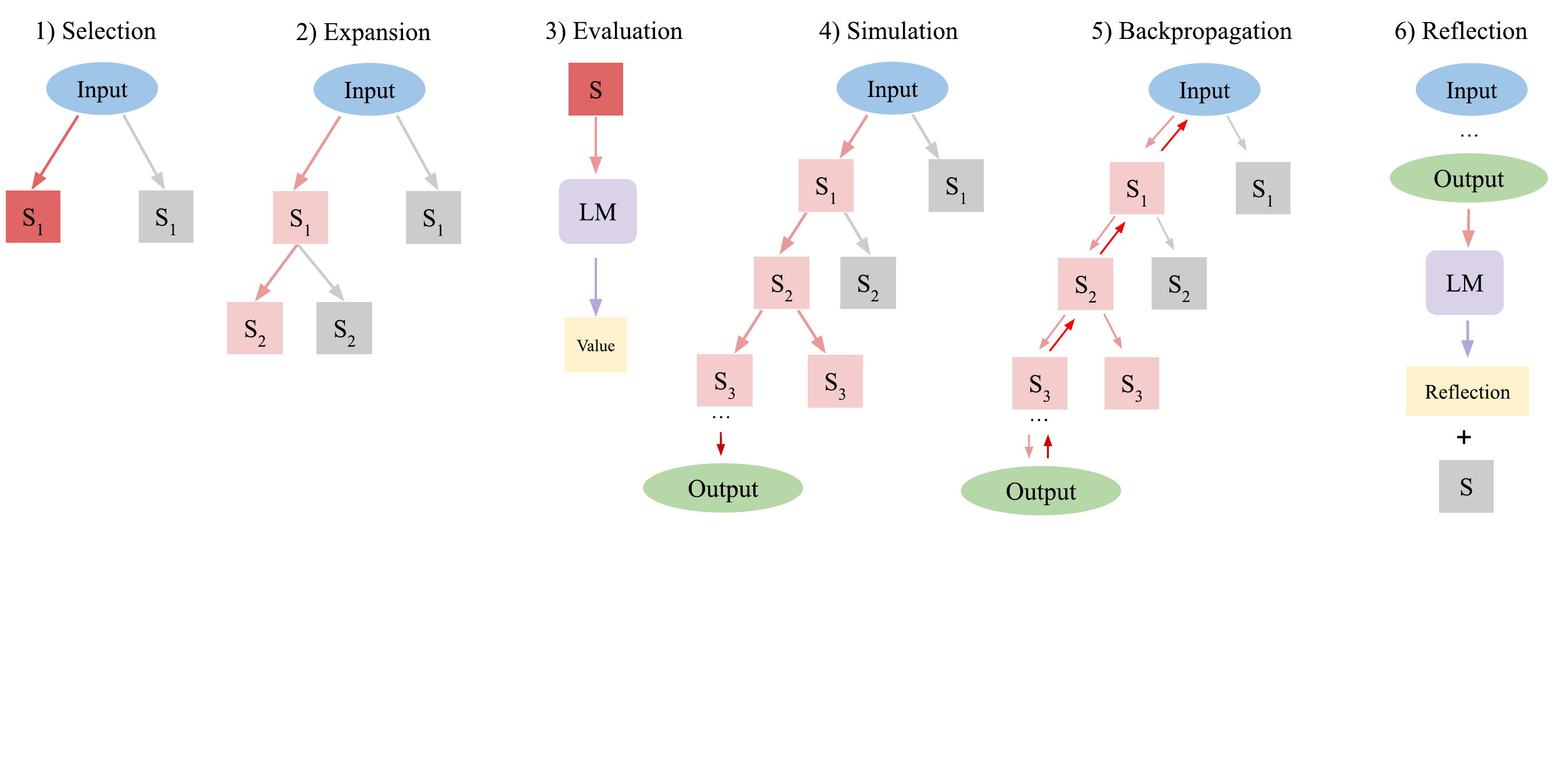}
    \vspace{-1.2in}
    \caption{Overview of the six operations in LATS. A node is \textit{selected}, \textit{expanded}, \textit{evaluated}, then \textit{simulated} until a terminal node is reached, and then the resulting value is \textit{backpropagated}. If the trajectory fails, a \textit{reflection} is generated and used as additional context for future trials. These operations are performed in succession until the budget is reached or the task is successful.}
    \vspace{-0.1in}
    \label{fig:phases}
\end{figure*}

The main component of LATS is a search algorithm that controls the problem-solving process with planning. To find the most promising trajectory and systemically balance exploration with exploitation, we adopt a variant of MCTS that frames decision-making as a tree search, in which each node $s=[x, a_{1\cdots i}, o_{1\cdots i}]$ represents a state comprising the original input $x$, action sequence $a_{1\cdot i}$, and observation sequence $o_{1\cdot i}$, where $i$ is a token in the text sequence. 

Our main technical contribution is \textit{adapting MCTS to language agents}. LATS repurposes $p_\theta$ as an agent, state evaluator, and feedback generator, leveraging the useful language representations of modern LMs to facilitate planning. While standard MCTS and RAP \cite{hao2023reasoning} rely on internal dynamics models to facilitate simulation, LATS uses environment interaction and does not require a world model. As depicted in Fig.~\ref{fig:phases}, LATS consists of a series of operations -- \textit{selection, expansion, evaluation, simulation, backpropagation, and reflection} -- performed in succession until the task is successfully completed or a computational limit is reached after sampling $k$ trajectories. The full pseudocode of LATS can be found in Sec.~\ref{sec:pseudocode} in the Appendix.

\textbf{Selection.}
In the first operation, the algorithm identifies a segment of the current tree most suitable for subsequent expansion. Starting from the root node, denoted as the initial state \(s_0\), a child node is selected at each tree level until a leaf node is reached. To balance exploration and exploitation, we use the UCT algorithm as shown in Eq.~\ref{eq:uct}. 

\vspace{0.5mm}
\textbf{Expansion.} 
After selecting a node, the second operation expands the tree by sampling $n$ actions from $p_\theta$, as described in the prior section. The environment receives each action and returns corresponding feedback as an observation. This results in $n$ new child nodes added to the tree. This tree is stored in an external long-term memory structure.

\vspace{0.5mm}
\textbf{Evaluation.} 
The third operation assigns a scalar value to each new child node for selection and backpropagation. This value effectively quantifies the agent's progress in task completion, serving as a heuristic to steer the search algorithm towards the most promising regions of the tree. As LATS does not involve training, we propose a novel value function for this setting based on two components: (1) a \emph{self-generated} LM score and (2) a \emph{self-consistency} score.

Inspired by ToT, we repurpose $p_\theta$ into a value function by prompting it to reason about a given state. To obtain a scalar value, we instruct $p_\theta$ to end its reasoning trace with a score indicating the correctness of the trajectory. Our key distinction from ToT is that we obtain this value after obtaining the environmental feedback, improving value assignment. This also enables scaling to more challenging environments, as it is difficult for LMs to improve their responses without external feedback \cite{huang2023large}. Additionally, to further improve value assignment, we introduce an additional heuristic based on self-consistency \cite{wang2022self}, in which actions sampled multiple times at the same state tend to be more accurate. This results in the overall value function:
\begin{equation}
\small
    V(s) = \lambda * \text{LM}(s) + (1-\lambda) * \text{SC}(s),
\end{equation}
where $\lambda$ is a hyperparameter. Notably, our method offers enhanced flexibility over programmed heuristics \citep{campbell2002deep} and greater efficiency than learned heuristics \citep{silver2017mastering}.

\textbf{Simulation.} The fourth operation expands the currently selected node until a terminal state is reached. At each depth level, we sample and evaluate nodes with the same operations but prioritize nodes of the highest value. Reaching a terminal state provides objective feedback on the correctness of a trajectory. If the task is completed successfully, then LATS terminates the search. If the solution is partially successful or unsuccessful, then we perform two additional operations as described below. The success of a trajectory is determined by the design of the specific environment, such as finalizing a purchase in web navigation environments.

\textbf{Backpropagation.} 
This operation updates the values of the tree based on the outcome of a trajectory. For each node $s_0,s_1,\dots, s_l$ in the trajectory from root (initial state $s_0$) of the searching tree to leaf (terminal state $s_l$), its value is updated to reflect the outcome of the simulation by $N(s_i)=N(s_{i-1})+1$ and $V(s_i)=\frac{V(s_{i-1})N(s_{i-1})+r}{N(s_i)}$, where $r$ is the reward. These updated values are used in the UCT formula (Eq.~\ref{eq:uct}) to guide the selection of the next node.

\textbf{Reflection.}
In addition to the environmental feedback, we leverage \textit{self-reflection} to further refine the decision-making process \citep{shinn2023reflexion, madaan2023selfrefine}. Upon encountering an unsuccessful terminal node, $p_\theta$ is prompted with the trajectory and final reward to provide a verbal self-reflection that summarizes the errors in the reasoning or acting process and proposes superior alternatives. We store both failed trajectories and corresponding reflections in the memory. In subsequent iterations, these are integrated as additional context to the agent and value function, refining both through in-context learning. This imparts a semantic gradient signal more useful than a scalar value, enabling the agent to learn from trial and error without the cost of expensive optimization such as reinforcement learning.

\textbf{Discussion.} Conceptually, LATS has several notable advantages as a general framework for reasoning and decision-making with LM agents.
(1) \textit{Generality}: LATS supports both reasoning and decision-making tasks by defining a shared space of thoughts and actions. (2) \textit{Deliberation}:  Leveraging MCTS and LM value function in LATS ensures a principled search that selects options with high value while exploring promising alternatives. (3) \textit{Adaptability}: Incorporating external feedback through observations and self-reflection in LATS enables greater adaptation during problem-solving. (4) \textit{Flexibility}: LATS can accommodate different scenarios, environments, and resource stipulations by modifying state design and tree dimensions. (5) \textit{Modularity}: The base LM agent, reflection generator, and value function can be independently altered and adapted to individual LM properties.

%around the use of the following The use of LATS is designed to

\begin{table}[t]
    \centering
    \begin{tabular}{l|c}
    \toprule
        \textbf{Prompt Method} & \textbf{HotpotQA (EM) $\uparrow$} \\
    \midrule
        Base LM & 0.32 \\
        CoT {\scriptsize~\citep{wei2022chain}} & 0.34 \\ 
        CoT - SC {\scriptsize~\citep{wang2022self}} & 0.38 \\
        ToT {\scriptsize~\citep{yao2023tree}} & 0.55 \\
        RAP {\scriptsize~\citep{hao2023reasoning}} & 0.60 \\
        RAP ($n = 10$) & 0.60 \\
        LATS (CoT) & \textbf{0.62} \\
    \bottomrule
    \end{tabular}
    \caption{GPT-3.5 \emph{reasoning}-based prompting results on HotpotQA. LATS achieves the highest exact match (EM) for reasoning. We sample $n=5$ nodes during expansion and $k = 50$ trajectories.}
    \label{table:hotpot-cot}
\end{table}
%~\yxw{we did not explain what is 1/O}\kai{explained}

% \begin{table}[t]
%     \centering
%     \begin{tabular}{l|c}
%     \toprule
%         \textbf{Prompt Method} & \textbf{HotpotQA (EM)} \\
%     \midrule
%         I/O & 0.32 \\
%         CoT {\scriptsize~\citep{wei2022chain}} & 0.34 \\ 
%         CoT - SC {\scriptsize~\citep{wang2022self}} & 0.38 \\
%         ToT {\scriptsize~\citep{yao2023tree}} & 0.55 \\
%         RAP (n = 5){\scriptsize~\citep{hao2023reasoning}} & 0.60 \\
%         RAP (n = 10) & 0.60 \\
%     \midrule
%         ReAct {\scriptsize~\citep{yao2023react}} & 0.32 \\
%         ReAct (best of k) & 0.38 \\
%         Reflexion {\scriptsize~\citep{shinn2023reflexion}} & 0.51 \\
%         LATS (n = 3) & 0.56 \\
%         LATS (n = 5) & 0.61 \\
%         LATS (n = 10) & 0.64 \\
%         LATS (CoT + ReAct) & \textbf{0.71} \\
%     \midrule
%         Supervised SoTA {\scriptsize~\citep{zhu2021adaptive}} & 0.675 \\
%     \bottomrule
%     \end{tabular}
%     \caption{GPT-3.5 prompting results on HotpotQA. The table is separated into reasoning-based prompting, acting-based prompting, and supervised training. LATS achieves the highest success rate, surpassing internal reasoning, and is competitive to supervised training. \hwc{put this table a side one?}}
%     \label{table:hotpot}
% \end{table}

\section{Experiments}
\label{sec:exp}

To demonstrate the general applicability of LATS, we evaluate our method on a variety of domains that require reasoning and acting: programming \citep{chen2021evaluating, austin2021program}, HotPotQA \citep{yang2018hotpotqa}, WebShop \citep{yao2022webshop}, and Game of 24~\cite{yao2023tree}.
%~\yxw{did not mention game 24?}\kai{added}

\subsection{HotPotQA}
\label{sec:HotPotQA}

\begin{table}[t]
    \centering
    \begin{tabular}{l|c}
    \toprule
        \textbf{Prompt Method} & \textbf{HotpotQA (EM) $\uparrow$} \\
    \midrule
        ReAct {\scriptsize~\citep{yao2023react}} & 0.32 \\
        ReAct (best of $k$) & 0.38 \\
        Reflexion {\scriptsize~\citep{shinn2023reflexion}} & 0.51 \\
        \textit{ToT (ReAct)} & 0.39 \\
        \textit{RAP (ReAct)} & 0.54 \\ 
        LATS (ReAct) & 0.63 \\
        LATS ($n = 3$) & 0.58 \\
        LATS ($n = 10$) & 0.65 \\
        LATS (CoT + ReAct) & \textbf{0.71} \\
    \bottomrule
    \end{tabular}
    \caption{GPT-3.5 \emph{acting}-based prompting results on HotpotQA. LATS achieves the highest exact match (EM) for acting. We sample $n=5$ nodes and use $k = 50$ trajectories. We also evaluate sampling ReAct $k$ times and using both CoT and ReAct base prompting designs for LATS, which achieves the best performance. Note that LATS outperforms ToT and RAP with ReAct prompting, which are the simple adaptations of search algorithms to decision-making.}
    \label{tab:hotpot-react}
\end{table}

For a task that can be approached with both reasoning-based and acting-based strategies, we consider HotPotQA \citep{yang2018hotpotqa}, a multi-hop question-answering benchmark that requires retrieval over two or more Wikipedia passages. For the action space, in addition to LM thoughts, we follow the setup from \citet{yao2023react}, which provides the agent with API calls to search and retrieve information. The output of these API calls and self-generated reflections form the observation space. Note that consistent with previous work \cite{yao2023react, shinn2023reflexion}, we use an oracle setup for HotPotQA, in which the environment provides feedback about the answer's correctness upon receiving an answer. This enables a fair comparison between our method and baselines in scenarios where the quality of feedback is high, allowing us to focus our evaluation on how well the agent incorporates external feedback. We use a subset of 100 questions and three few-shot examples for each method. For ToT, we use DFS as the base search algorithm. For all methods that involve sampling, including LATS, we sample $k=50$ trajectories. More details are in Appendix Sec.~\ref{sec:envdetail}.
%~\yxw{this sentence is written in a confusing way - it sounds like we have two settings, with an oracle one and another one, but it should be only the oracle one? I guess the confusion lies in using `additionally' to start the sentence, maybe change to `Note that'? also, need to emphasize this is for all the methods including baselines.}\kai{Andy - please deal with this}\az{added}}

We evaluate internal reasoning strategies by removing actions and observations from the context, corresponding to CoT \citep{wei2022chain} and its variants, CoT-SC \citep{wang2022self}, ToT \citep{yao2023tree}, and RAP \citep{hao2023reasoning}. These methods rely solely on the agent's existing knowledge to answer the question. We further consider acting-based methods ReAct, Reflexion, and LATS, which augment the agent with the interactive API environment and primarily evaluate its information retrieval abilities. We also design a simple integration of search algorithms with LM agents, extending ToT and RAP with ReAct prompting to handle external observations. In addition, while LATS is designed for scenarios where external feedback can enhance reasoning, we also implement a reasoning-only version with CoT as the base prompting framework. Moreover, we combine internal and external reasoning in LATS by first prompting with a CoT-based prompt and then switching to a ReAct-based prompt upon failure. This is closer to how humans might approach this task by using tools to retrieve additional information only when the answer is not already known. 

\textbf{Results.} We observe in Tab.~\ref{table:hotpot-cot} and Tab.~\ref{tab:hotpot-react} that both internal reasoning and external retrieval strategies perform well on HotPotQA. Due to their large-scale training corpus, modern LMs already encode factual knowledge and can often directly answer the question correctly. While CoT can slightly enhance performance on questions requiring reasoning, larger gains are observed with search methods ToT and RAP (Tab.~\ref{table:hotpot-cot}, Row 4, 5), which can sample and explore more outputs. We observe similar results for acting-based methods. LATS surpasses ReAct, even when sampling the same number of trajectories, by expanding more nodes with principled search. This is demonstrated when modifying $n$, the number of nodes expanded during each iteration. Increasing $n$ can consistently improve performance, although at greater computational and inference costs. LATS also outperforms RAP on internal reasoning, but has higher performance on the decision-making setting of HotPotQA than the reasoning setting. Contrary to LATS, the ReAct versions of ToT and RAP (Tab.~\ref{tab:hotpot-react}, Row 4, 5) \textit{perform even worse than the reasoning-only setting} of HotPotQA, which indicates that the acting-based setting is more challenging and \emph{adaptation of search algorithms to decision-making scenarios is non-trivial}. Combining internal and external reasoning in LATS results in the highest performance, indicating the importance of external feedback in augmenting reasoning even in tasks where the base LM can already perform.
%~\yxw{worse than?- this sentence is confusing}\kai{Andy - please clarify this}\az{added}.

\subsection{Programming}

\begin{table}[t]
    \centering
    \begin{tabular}{ll | c}
    \toprule
    \textbf{Prompt Method} & \textbf{Model} & \textbf{Pass@1 $\uparrow$} \\
    \midrule
    CoT {\scriptsize~\citep{wei2022chain}} & GPT-3.5 & 46.9 \\
    ReAct {\scriptsize~\citep{yao2023react}} & GPT-3.5 & 56.9 \\
    Reflexion {\scriptsize~\citep{shinn2023reflexion}} & GPT-3.5 & 68.1 \\
    ToT {\scriptsize~\citep{yao2023tree}} & GPT-3.5 & 54.4 \\
    RAP {\scriptsize~\citep{hao2023reasoning}} & GPT-3.5 & 63.1 \\
    LATS (ReAct) & GPT-3.5 & \textbf{83.8} \\ 
    \midrule
    Base LM & GPT-4 & 80.1 \\
    Reflexion & GPT-4 & 91.0 \\
    LATS (ReAct) & GPT-4 & \textbf{92.7} \\
    \bottomrule
    \end{tabular}
    \caption{GPT-3.5 and GPT-4 Pass@1 accuracy on HumanEval. Prompting with LATS achieves the best performance. We sample 5 solutions during expansion for 8 iterations.}
    \label{tab:humaneval}
\end{table}

To demonstrate the importance of external observations for complex reasoning tasks, we evaluate the baselines and LATS on programming with HumanEval~\citep{chen2021evaluating}\footnote{Some baselines use 161 questions from HumanEval. We use all 164 questions for LATS and find minimal performance differences, so we report baselines for both settings.} and MBPP~\citep{austin2021program}. Both datasets measure the correctness of synthesized programs in Python from natural language docstrings. We use individual solutions as the action space and test suite and compiler feedback as the external observation. We follow \citet{bei2022codet} and use an LM to generate a synthetic test suite of syntactically valid ``assert'' statements for each question. For each step, the solution is evaluated on this test suite, and the results, including successful and failed tests and compiler output, are added to the context as an observation. 

For this task, the reasoning and acting baselines share an action space, but acting methods are able to incorporate observations as additional context. For LATS, since each action corresponds to a complete solution, we skip the simulation step of LATS and directly use the percentage of passed tests as the backpropagated reward. We use $k = 8$ iterations, set the number of generated tests at $4$, and sample $n = 5$ solutions during expansion. After the search is completed, we select the solution with the highest value and evaluate it on the real test suite for the pass@1 accuracy evaluation. More details can be found in Appendix Sec.~\ref{sec:envdetail}.

\textbf{Results.} Tab.~\ref{tab:humaneval} and Tab.~\ref{table:mbpp} show that both search and semantic feedback are crucial for better performance. Despite not using observations, ToT and RAP are competitive with Reflexion. LATS has the highest performance on both datasets. RAP uses a search algorithm similar to LATS, which reveals the importance of external feedback for difficult reasoning tasks such as programming. With GPT-4, using LATS sets the state of the art for HumanEval, validating that LATS can be used with more advanced LMs for higher performance.

\begin{table}[t]
    \centering
    \begin{tabular}{l | c}
    \toprule
    \textbf{Prompt Method} & \textbf{Pass@1 $\uparrow$} \\
    \midrule
    CoT {\scriptsize~\citep{wei2022chain}} & 54.9 \\
    ReAct {\scriptsize~\citep{wei2022chain}} & 67.0 \\
    Reflexion {\scriptsize~\citep{shinn2023reflexion}} & 70.0 \\
    ToT {\scriptsize~\citep{yao2023tree}} & 65.8 \\
    RAP {\scriptsize~\citep{hao2023reasoning}} & 71.4 \\
    LATS (ReAct) & \textbf{81.1} \\
    \bottomrule
    \end{tabular}
    \caption{GPT-3.5 Pass@1 accuracy on MBPP. Prompting with LATS achieves the highest performance. We sample 5 solutions during expansion for 8 iterations.}
    \label{table:mbpp}
\end{table}

\subsection{WebShop}

For a complex decision-making environment with practical applications, we consider WebShop \citep{yao2022webshop}, an online shopping environment composed of a website with 1.18M real-world products and 12k human instructions. Agents must navigate a website through a variety of commands to purchase an item matching a user specification. We use the preconstructed action space of search and click commands and browser feedback and reflections for the observation. The performance is gauged using two metrics: an average score, reflecting the percentage of user-specified attributes met by the selected product, and a success rate, indicating the frequency with which the chosen product fulfills all given conditions. We compare against acting-based prompting methods and RL-based approaches. We evaluate on 50 instructions, expand $n=5$ children for LATS, and set $k=30$ for LATS, ReAct (best of $k$), and Reflexion. More details and prompts are in Appendix Sec.~\ref{sec:envdetail} and Sec.~\ref{sec:webshop_prompt}.

\textbf{Results.} We find in Tab.~\ref{table:webshop} that GPT-3.5 with ReAct is competitive to imitation learning (IL) and can exceed reinforcement learning techniques with stronger prompting strategies. Sampling $k=30$ trajectories with ReAct and Reflexion results in a similar performance, suggesting the semantic feedback is not as helpful in complex environments like WebShop. Similar to \citet{shinn2023reflexion}, we find that generated reflections are often generic and do not provide useful feedback, resulting in a tendency for the agent to become stuck in local minima. However, using LATS indeed results in a noticeable improvement, indicating a more effective exploration for the same number of iterations.

\begin{table}[t]
  \centering
    \begin{tabular}{c | cc}
    \toprule
    Method & Score $\uparrow$ & SR $\uparrow$ \\
    \midrule
    ReAct {\scriptsize~\citep{yao2023react}} & 53.8 & 28.0 \\
    ReAct (best of k) & 59.1 & 32.0 \\
    Reflexion {\scriptsize~\citep{shinn2023reflexion}} & 64.2 & 35.0 \\
    LATS (ReAct) & \textbf{75.9} & \textbf{38.0} \\
    \midrule
    IL{\scriptsize~\citep{yao2022webshop}} & 59.9 & 29.1 \\
    IL+RL{\scriptsize~\citep{yao2022webshop}} & 62.4 & 28.7 \\
    Fine-tuning {\scriptsize~\citep{furuta2023multimodal}} & 67.5 & 45.0 \\
    \midrule
    \textit{Expert} & 82.1 & 59.6 \\
    \bottomrule
    \end{tabular}
    \caption{
      Score and success rate (SR) on WebShop. Results are organized into prompting, RL-based training, and human performance. For the same number of iterations, LATS improves both score and SR and surpasses RL-based training.
    }
    \label{table:webshop}
\end{table}

\begin{table}[t]
  \centering
  \begin{tabular}{l|c}
  \toprule
    Prompt Method & Game of 24 (Success Rate) $\uparrow$ \\
    \midrule
    CoT {\scriptsize~\citep{wei2022chain}} & 0.08 \\
    Reflexion {\scriptsize~\citep{shinn2023reflexion}} & 0.12 \\ 
    ToT {\scriptsize~\citep{yao2023tree}} & 0.20 \\
    RAP {\scriptsize~\citep{hao2023reasoning}}& 0.40 \\
    % LATS ($\lambda=1$) & 0.40  \\
    LATS (CoT) & \textbf{0.44}  \\
    \bottomrule
    \end{tabular}
    \caption{Results on Game of 24 with GPT-3.5. We sample $n=5$ nodes and $k=30$ trajectories. % For our proposed value function, we use $\lambda = 0.5$, which allows LATS to outperform previous methods proposed for reasoning tasks.
    }
    \label{table:game24}
\end{table}

\subsection{Ablation Study and Additional Analysis}%Additional Observations}

We further test the reasoning ability of LATS on Game of 24, and also conduct additional experiments on HotPotQA to demonstrate the effect of each component of LATS (results shown in Tab.~\ref{table:abl}). % We use HotPotQA as our setup incorporates both reasoning (through thoughts) and acting (through API calls);  
More ablations for token consumption on HotPotQA are in Tab.~\ref{tab:cost} in Appendix Sec.~\ref{sec:abl}. 
%In this section, w

\textbf{Reasoning on Game of 24.} To show how LATS can be applied to purely internal reasoning tasks, we additionally evaluate on Game of 24 \cite{yao2023tree}, a mathematical reasoning task where the agent must construct 24 out of a set of numbers and basic operations. We use CoT as the base prompting design and employ the same operations as in other settings. We find in Tab.~\ref{table:game24} that LATS outperforms previous methods proposed specifically for reasoning. This is due to our proposed value function, which incorporates self-consistency as an additional heuristic. 

\begin{table}[t]
    \centering
    \begin{tabular}{l | c}
    \toprule
    Prompt Method & HotPotQA (EM) $\uparrow$ \\
    \midrule
    ToT (ReAct) & 0.39 \\
    RAP (ReAct) & 0.54 \\ 
    LATS (No LM Heuristic) & 0.37  \\
    LATS (DFS) & 0.42  \\
    LATS (No Reflection) & 0.58  \\
    LATS (ReAct) & \textbf{0.63} \\
    \bottomrule
    \end{tabular}
    \caption{Ablation results on LATS and baseline variants in HotPotQA. We use ReAct as the base prompt and sample $n=5$ children and $k=50$ trajectories. LATS requires every component and operation for optimal performance.}
    \label{table:abl}
\end{table}

\textbf{Self-reflection.} LATS uses self-reflection to provide additional semantic signals for the agent. In Tab.~\ref{table:abl} (Row 5, 6), we observe a $0.05$ performance drop when self-reflection is removed from LATS, validating its usefulness. This is a smaller gain than the $0.19$ gain that Reflexion has over ReAct as shown in Tab.~\ref{tab:hotpot-react}, suggesting overlap between the questions where an answer can be improved by self-reflection and search. This variant outperforms RAP (ReAct), reflecting our improvements to MCTS.

\begin{table*}[t]
    \small
    \centering
    \begin{tabular}{c|c|c|c}
        \toprule
        Method & Performance $\uparrow$ & Sample complexity $\downarrow$ & Token Consumption $\downarrow$ \\
        \hline
         ReAct (Best $k=250$) & $0.42$ & $O(k)$ & - \\
         CoT-SC ($n=1, k=250$) & $0.40$ & $O(k)$ & -  \\
         LATS ($n=1, k=50$) & $0.48$ & $O(k)$ & -  \\
         ToT (ReAct, $n=5,k=50$) & $0.49$ & $O(kn)$ & $210,215$\\
         RAP (ReAct, $n=5,k=50$) & $0.54$ & $O(kn)$ & $176,500$\\
         LATS ($n=5,k=50$) & $0.63$ & $O(kn)$ & $173,290$\\
         \bottomrule
    \end{tabular}
    \caption{Performance, sample complexity of different methods, average number of nodes expanded, and token consumption upon success by methods with tree-based search. $n$ is the number of children nodes expanded at every step and $k$ is the number of trajectories. LATS has the same sample complexity as other methods with tree-based search and expands less nodes upon success, which indicates lower token cost.}
    \label{tab:cost}
\end{table*}

\begin{table}[t]
\centering
\begin{tabular}{l|ccc}
\toprule
Method & $k$ & HotPotQA $\uparrow$ & \# of Nodes $\downarrow$ \\
\midrule
ToT & 10 & 0.34 & 33.97 \\
RAP & 10 & 0.44 & 31.53 \\
LATS & 10 & 0.44 & 28.42 \\
\midrule
ToT & 30 & 0.39 & 47.54 \\
RAP & 30 & 0.50 & 37.71 \\
LATS & 30 & 0.52 & 34.12 \\
\midrule
ToT & 50 & 0.49 & 84.05 \\
RAP & 50 & 0.54 & 70.60 \\
LATS & 50 & \textbf{0.61} & \textbf{66.65} \\
\bottomrule
\end{tabular}
\caption{Comparison of the cost of different methods on HotPotQA. LATS achieves the highest accuracy and the lowest average number of nodes/states required for success at various $k$ trajectories sampled.}
\vspace{-0.1in}
\label{table:costk}
\end{table}

\textbf{Search algorithm.} MCTS is a more principled search algorithm than variants like A* \cite{zhuang2023toolchain} or DFS and is the basis for observed performance gains. We observe the effects of using DFS, and incorporate the LM-based heuristic used in ToT in which branches with low values are pruned. This removes the selection and backpropagation operations, and we observe a $0.21$ drop in performance in Tab.~\ref{table:abl} (Row 4) when sampling the same number of nodes but outperforms ToT (ReAct). {Despite also benefiting from ground-truth feedback, LATS uses it better than ToT and RAP and can outperform these methods.} We also find in Tab.~\ref{table:abl} (Row 3) that LM scoring, the main component of our value function, is crucial for leveraging external feedback and strong performance.

{\textbf{Sample complexity and token consumption.} One possible concern of LATS is that the tree-structured search might consume much more tokens than existing methods. To further study the computational cost of LATS compared to prior methods, we examine the sample complexity (i.e., asymptotic token cost) of all methods considered in this paper and count the average number of nodes expanded by our method and other tree-structured methods (ToT and RAP) upon successful search on HotPotQA. We present the results in Tab.~\ref{tab:cost} and Tab.~\ref{table:costk}, which show that our method has the same sample complexity as other tree-based search methods and requires fewer overall tokens and states. The token cost gap will be even larger when taking failed trajectories into account, since our method has a higher success rate and reaches the computational budget limit less often. This is also true when sampling a smaller number of trajectories; on average, LATS requires 3.55 fewer nodes than RAP and 12.12 fewer nodes than ToT. These findings underscore our improvements to MCTS and adaptation to LM agents, resulting in a more principled and efficient search mechanism.}

\vspace{0.2mm}
\section{Conclusion}
\label{sec:conclusion}
\vspace{1.2mm}

This work introduces Language Agent Tree Search (LATS), the first framework to unify reasoning, acting, and planning for enhanced LM problem-solving. LATS addresses key limitations of prior prompting techniques by deliberately constructing trajectories with search algorithms, incorporating external feedback, and enabling agents to learn from experience. Our evaluation demonstrates the ability of LATS to harness LM capabilities for various decision-making tasks while maintaining its reasoning ability \emph{without additional training}. The proposed synergies between search, interaction, and reflection offer a versatile approach to autonomous decision-making, highlighting the potential of LMs as generalist agents. 

\vspace{1.2mm}
{\textbf{Limitations and future directions.} LATS has two main limitations that should be considered before its application. First, it has a higher computational cost compared to simpler prompting methods like ReAct or Reflexion, which may limit its practicality in certain situations. Second, LATS assumes the ability to revert to earlier states in decision-making environments, which may not be universally applicable in all possible environments. Despite these limitations, it is worth noting that LATS still achieves better performance and efficiency compared to similar methods, and the number of nodes expanded at each step provides a trade-off between performance and efficiency. Additionally, we expect inference-time compute costs to decrease over time, thereby increasing the usefulness of LATS and other ``System-2'' LM approaches. Finally, the reversion property is feasible in many real-world applications, opening up new opportunities in the LM decision-making community. Future directions include scaling LATS to more complex environments or multi-agent frameworks and improving efficiency to reduce costs. A more detailed discussion about the limitations of LATS can be found in Appendix Sec.~\ref{sec:discussion}.}

% {\color{red}\textbf{Limitations.} Despite its impressive reasoning and decision-making ability on many tasks of different types, LATS has two limitations: 1) relative higher computational cost to simpler prompting methods like ReAct, and 2) the assumption that the environment could back up for decision-making tasks. See detailed discussion in Appendix~\ref{sec:discussion}.}
% Although LATS can improve reasoning and decision-making, this arrives at a higher computational cost relative to simpler prompting methods like ReAct or Reflexion. A full analysis of the cost can be found in Tab.~\ref{tab:cost} in Appendix~\ref{sec:abl}. While such an issue is mitigated by the fact that the number of nodes $n$ expanded at every step provides a natural trade-off between performance and efficiency (setting $n=1$ makes the method as efficient as ReAct with multiple trials or CoT-SC), in practice, we recommend using LATS for difficult tasks like programming or for situations where performance is prioritized over efficiency. We hope that continued advancements in LLMs will reduce costs and increase the practicality of LATS.

% harness LM capabilities for a variety of reasoning and decision-making tasks without additional training.
\newpage
\section*{Impact Statement}

LATS is a framework that enhances LM performance through interactions with an environment. This improvement in autonomous decision-making may facilitate harmful uses of LMs. On the other hand, LATS enhances interpretability and the potential for greater alignment, as it involves high-level linguistic reasoning and actions through several rounds of decision-making and reflection rather than relying on autoregressive generation. Finally, enhancing the capabilities of LM agents may raise security risks, such as executing malware. We encourage further research to fully understand and mitigate the risks of LMs.

\section*{Acknowledgements}

We thank Daniel Campos for useful feedback on earlier versions of this paper. This work was supported in part by NSF Grant 2106825, NIFA Award 2020-67021-32799, the Jump ARCHES endowment through the Health Care Engineering Systems Center at Illinois and the OSF Foundation, and the IBM-Illinois Discovery Accelerator Institute. This work used NVIDIA GPUs at NCSA Delta through allocations CIS220014, CIS230012, and CIS230218 from the ACCESS program.

\bibliography{main}
\bibliographystyle{plainnat}

\clearpage

\section*{Appendix of LATS}
\appendix

The appendix is organized as follows. First in Sec.~\ref{sec:pseudocode}, we show the pseudocode of our proposed algorithm, LATS. In Sec.~\ref{sec:discussion}, we provide further discussion of {the limitations of our method}. In Sec.~\ref{sec:abl}, we present additional experimental results. In Sec.~\ref{sec:envdetail}, we specify the environment details in our experiments. Finally, we list our prompts used for the three environments in Sec.~\ref{sec:hotpotqa_prompt} (HotPotQA), Sec.~\ref{sec:programming_prompt} (Programming), and Sec.~\ref{sec:webshop_prompt} (WebShop), respectively.

\section{LATS Pseudocode}
\label{sec:pseudocode}
Alg.~\ref{alg:mcts} shows the pseudocode of our algorithm LATS. Nodes are stored explicitly in the memory. Unless otherwise specified, in all experiments, we set the number of sampled nodes to $n=5$ and the exploration weight to $w=1$. We use a self-consistency weight of $\lambda=0.5$ for HotPotQA and Game of 24, and $\lambda=0.8$ for Programming and WebShop.
\begin{algorithm*}[ht]
\centering
\caption{$\operatorname{LATS}(s, p_\theta, {p_V}, p_{\text{ref}}, d, k, n, w, a, b)$ \label{alg:mcts}}
\begin{minipage}{0.9\linewidth} 
\small
\begin{algorithmic}
    \Require Initial state $s$, action generator $p_\theta$, value function {$p_V$}, reflection generator $p_{\text{ref}}$, number of generated actions $n$, depth limit $L$, number of roll-outs $K$, context $c$, exploration weight $w$, and value function weight $\lambda$
    \State Initialize action space $A$, observation space $O$
    \State Initialize the state-action value function ${p_V} : S \times A \mapsto \mathbb R$ and visit counter ${N} : S \mapsto \mathbb N$ to one
    \For {$k \gets 0, \dots, K - 1$}
        \For {$t \gets 0, \dots, L - 1$}
            \If {$s_t$ not\ terminal} \Comment{Expansion \& Simulation}
                \For {$i \gets 1, \dots, n$}
                \State Sample $a_t^{(i)} \sim p_\theta(s_t)$
                \State Get $o_t^{(i)}$ from environment, $s_{t+1}^{(i)} \gets (c_{t}^{(i)}, o_t^{(i)}, a_t^{(i)})$, $c_{t+1}^{(i)} \gets (o_t^{(i)}, a_t^{(i)})$
                \State Evaluate ${V}_t^{(i)} \sim \lambda * {p_V}(s_t^{(i)}) + (1-\lambda) * \text{SC}(s_t^{(i)})$ 
                \Comment Evaluation
                \State ${V}(s_t) \gets {V}_t^{(i)}$
                \State Add $s_t^{(i)}$ to children
                \EndFor
            \EndIf
            \If {$s_t$ is terminal}
                \Comment{Reflection}
                \State Get $r$ from environment
                \If {$r$ not success}
                    \State $\text{reflection} \gets p_{\text{ref}}(c_t)$
                    \State $c \gets \text{reflection}$
                \EndIf
            \EndIf
            \State $a_{t} \gets \arg\max_{a \in e(s_t)} \left[ {V(s_t)} + w \sqrt{\frac{\ln {N}(s_{t})}{{N}(s_{t+1})}} \right]$ \Comment{Selection}
            \State Get corresponding $o_t$ from memory, $s_{t+1} \gets (c_t,o_t,a_t), c_{t+1}\gets (o_t,a_t)$
            \State ${N}(s_{t+1}) \gets {N}(s_{t+1}) + 1$
            \State \algorithmicif\ {$a_t$ is an output action}\ \algorithmicthen\ \textbf{break}
        \EndFor
        \State $T \gets$ the actual number of steps
        \For {$t \gets T - 1, \dots, 0$} \Comment{Backpropagation}
            \State {$V(s_t)\leftarrow \frac{V(s_t)(N(s_t) - 1) + r}{N(s_t)}$}
        \EndFor
    \EndFor
\end{algorithmic}
\end{minipage}
\end{algorithm*}

\section{More Discussion on Limitations}

\label{sec:discussion}

As stated in Sec.~\ref{sec:conclusion}, LATS has two main limitations: 

\textbf{Computational cost.} Although LATS can improve reasoning and decision-making, this arrives at a higher computational cost relative to simpler prompting methods like ReAct or Reflexion. However, the following facts serve as mitigations to this issue:

\begin{itemize}
    \item Asymptotically, our method has the same sample complexity as ToT~\cite{yao2023tree} and RAP~\cite{hao2023reasoning}, but achieves better performance, expands fewer nodes, and uses fewer tokens on average upon success. This suggests that our method is not only stronger in problem-solving but also has higher efficiency. A full analysis of the cost can be found in Tab.~\ref{tab:cost} in Appendix~\ref{sec:abl}.
    \item The number of nodes $n$ expanded at every step provides a natural trade-off between performance and efficiency. In fact, setting $n=1$ makes the method as efficient as ReAct~\cite{yao2023react} with multiple trials or CoT-SC~\cite{wang2022self}.
\end{itemize}

In general, we recommend using LATS for difficult tasks like programming or for situations where performance is prioritized over efficiency in practice. We hope that continued advancements in LMs will reduce costs and increase the applicability of LATS.

{Additionally, there exists a minor cost from querying the environment, which we find to be trivial for the environments we study. Most LM-based environments involve API-based tools, which are inexpensive and fast to use. It is also worth noting that this is cheaper than the inference cost associated with using LMs as world models, as in previous search approaches \cite{hao2023reasoning, liu2023reason}.}

\textbf{Assumption of environment reversion in decision-making.} Since our method is based on Monte Carlo Tree Search and is model-free, one limitation of LATS on decision-making tasks is that it requires the agent to be able to revert to earlier states in the environments. However, this reversion property is feasible in many real-world environments and applications (despite being not universally applicable in all possible environments), including programming (HumanEval~\cite{chen2021evaluating}), web search (WebShop~\cite{yao2022webshop}), text-based manipulation tasks (Alfworld~\cite{shridhar2020alfworld}), and LMs with tool use (ToolBench~\cite{qin2023toolllm}). Therefore, we believe that leveraging the reversion property is not a shortcoming but rather \textit{a feature that has not been explicitly given notice by the LM decision-making community} -- it opens up new opportunities in the emerging LM agent community.

%However, since a variety of  LM task possesses such property, we feel that prior work overlooked and did not use this capability. Recognizing this fact and leveraging this property is the main motivation of our work.
%}
Additionally, the benchmarks we use in this paper are relatively simple and focused on decision-making compared to the complexity of real-world interactive environments. Moreover, some environments might not easily support rollbacks to previous states. However, the design of LATS is flexible and can be adjusted to various resource constraints. Using planning-based prompting methods like LATS in environments like Minecraft \citep{fan2022minedojo} and more reasoning benchmarks would be interesting avenues for future work. 

{
\section{Additional Ablations}\label{sec:abl}

\begin{figure}[t]
    \centering
    \begin{minipage}{.48\textwidth}
        \centering
        \begin{table}[H]
            \begin{tabular}{c|c}
                \toprule
                Prompt Method & HotpotQA (EM) $\uparrow$  \\
                \midrule
                LATS ($w=0.5$) & 0.55 \\
                LATS ($w=2.0$) & 0.63 \\
                LATS ($d=4$) & 0.58 \\
                LATS (CoT) & 0.62 \\
                LATS (No LM Heuristic) & 0.37 \\
                LATS ($w=1.0$, $d=7$) & \textbf{0.63} \\ 
                \bottomrule
            \end{tabular}
    \caption{Ablation results on LATS and baseline variants in HotPotQA measured by Exact Match (EM). We test different depth $d$, exploration factor $w$, and versions of LATS using CoT and without the LM value function. We sample $n=5$ and $k=50$ trajectories.}
        \end{table}

        \label{tab:hotpotqa}
    \end{minipage}%
    \hspace{4pt}
    \begin{minipage}{.48\textwidth}
        \centering
         \includegraphics[width=0.95\linewidth]{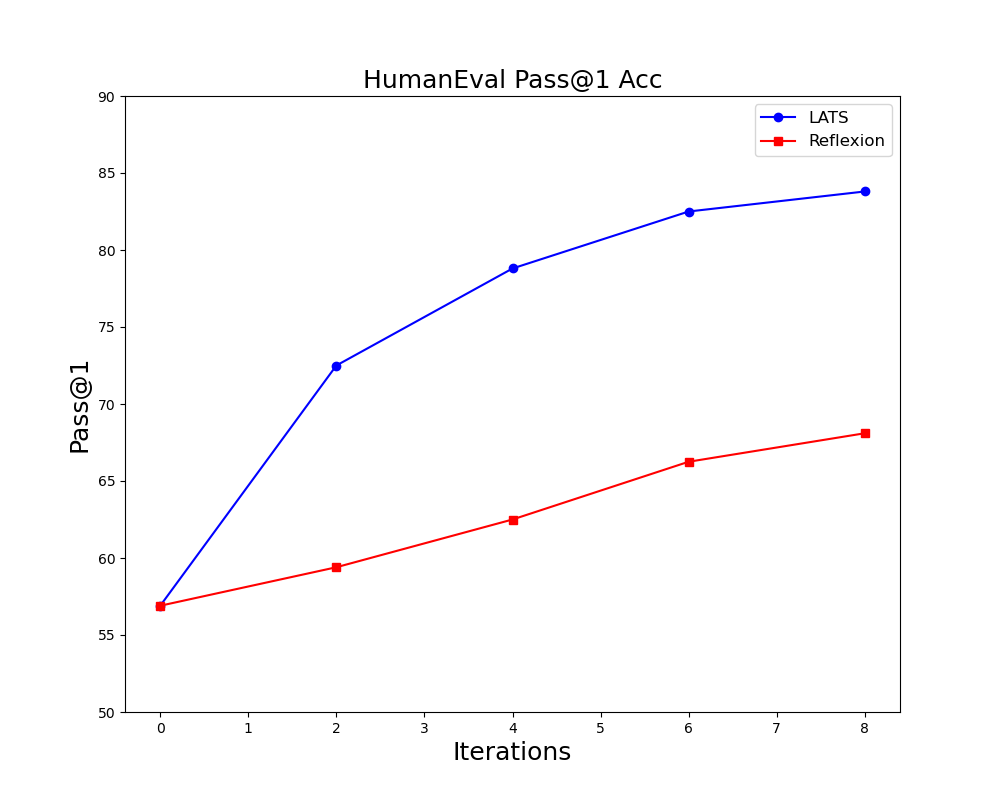}
        \vspace{-0.1in}
        \caption{Performance over successive iterations on HumanEval with GPT-3.5.}
        \label{fig:k}
    \end{minipage}
\end{figure}

In this section, we ablate various designs of LATS. Experiments are conducted on HotPotQA with a maximum of $k=50$ trajectories and sampling size of $n=5$ and HumanEval with a maximum of $k=8$ trajectories and sampling size of $n=5$. The result for HotPotQA is shown in Tab.~\ref{table:abl} and HumanEval in Fig.~\ref{fig:k}.

\textbf{Exploration weight.} We find that there is lower performance on HotPotQA when the exploration weight $w$ in the selection formula is decreased to $0.5$, suggesting that this reduces the effectiveness of the search. Increasing $w$ to $2.0$ does not lead to a performance improvement, but we tend to observe faster convergence. The optimal setting depends on the particular environment and complexity of the state space.

\textbf{Depth.} In our main experiments we use a maximum depth of $d=7$ on HotPotQA for all methods, following previous work \citep{yao2023react}. We ablate the effect on LATS after reducing it to $d=4$. This results in only a slight drop in performance. We find that most questions can be answered within four steps, and using a greater number of steps tends to force the agent into local minima and rarely improves success.

\textbf{LM value function.} The LM value function scores states based on expected future reward. Without this heuristic, the only signal to guide search would be from environment rewards for completed trajectories, which are scarce and often binary. When we remove the evaluation operation, we observe a dramatic $0.26$ drop in performance.

\textbf{Performance over time.} To see the effects of increasing the number of trajectories sampled, we change $k$ to different values. We conduct this experiment on HumanEval, which has a more noticeable difference due to sampling less trajectories. The results are shown in Fig. \ref{fig:k}, in which LATS scales better with more iterations than Reflexion.

\section{Environment Details}
\label{sec:envdetail}
\subsection{HotPotQA}

\begin{figure*}[t]
    \centering
    \includegraphics[width=1.0\linewidth]{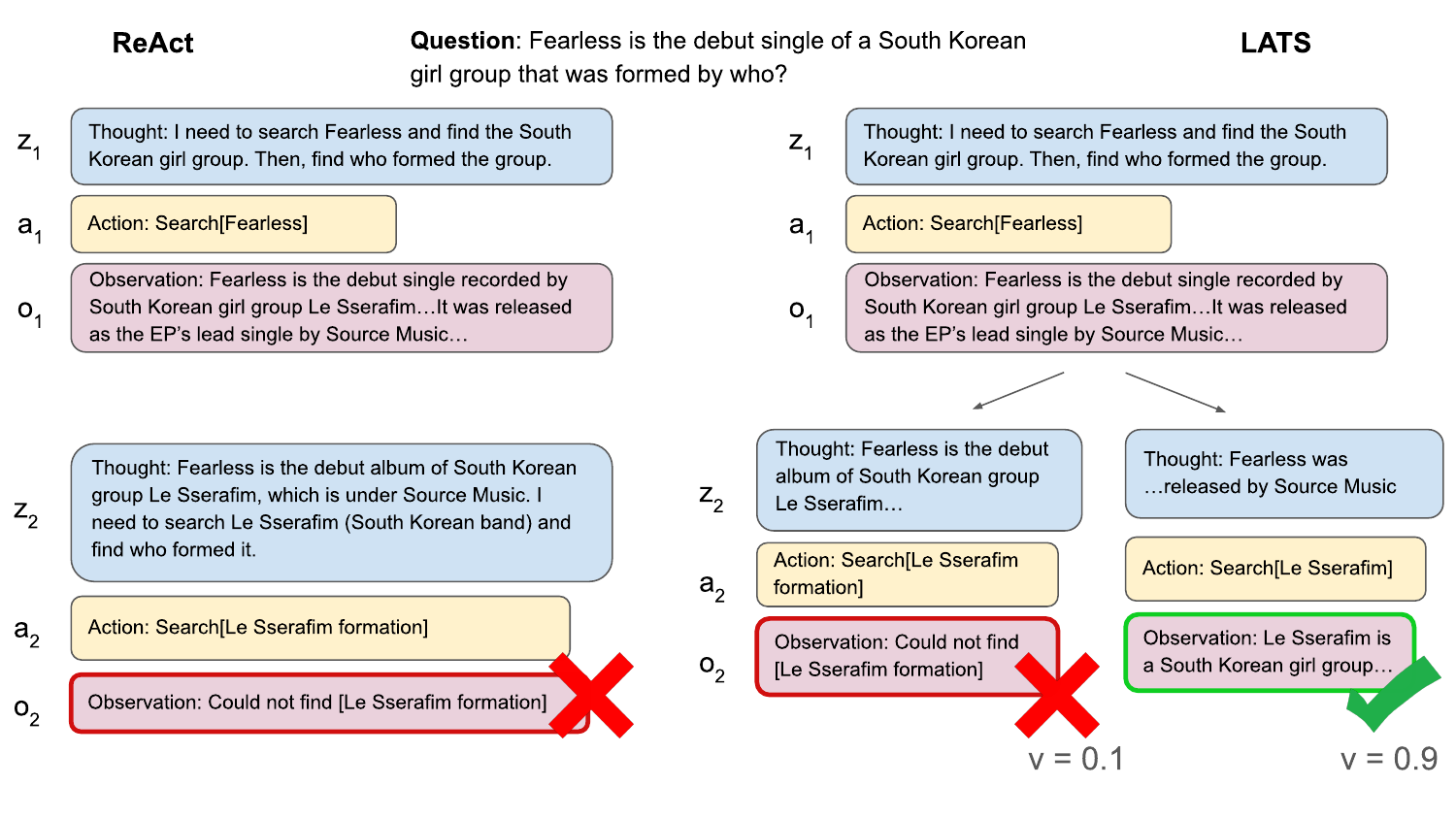}
    \caption{{Example trajectories on} HotPotQA for ReAct (\textbf{left}) and LATS (\textbf{right}). LATS can sample more actions and avoid failure from previous mistakes by evaluating states with an LM to guide the search toward promising areas of the tree.}
    \label{fig:qualitative}
    \vspace{-0.1in}
\end{figure*}

HotPotQA \citep{yang2018hotpotqa} is a question-answering dataset that requires reasoning over multiple supporting documents to answer questions. It contains 113k Wikipedia-based question-answer pairs crafted by crowdworkers to be diverse, multi-hop, and explainable. Questions cover a range of types like entities, locations, dates, and comparison of shared properties between two entities. Crowdworkers also provide supporting facts from the documents that justify the answer. We use the HotPotQA benchmark setting with all the Wikipedia paragraphs to test retrieval. We use a randomly selected subset of 100 questions for our experiments and a maximum depth limit of 6. {Fig.~\ref{fig:qualitative} illustrates how ReAct and LATS work on an example task of HotPotQA, and gives a qualitative example on how LATS outperforms ReAct on the task. } For value function hyperparameters, we use $\lambda = 0.5$ for the LM score and self-consistency score.

\textbf{Action Space.} We adopt the Wikipedia web API proposed in \citet{yao2023react}, with three types of actions to support interactive information retrieval: 

(1) \textbf{\texttt{search}}[\texttt{entity}], which returns the first 5 sentences from the corresponding \texttt{entity} wiki page if it exists, or else suggests top-5 similar entities from the Wikipedia search engine, 

(2) \textbf{\texttt{lookup}}[\texttt{string}], which returns the next sentence in the page containing \texttt{string}, 

(3) \textbf{\texttt{finish}}[\texttt{answer}], which finishes the current task with \texttt{answer}.

These API calls and free-form thoughts form the action space for this environment.

\subsection{Programming}

The HumanEval dataset \citep{chen2021evaluating} is a collection of 164 handwritten programming problems introduced to evaluate the functional correctness of models for synthesizing programs from natural language descriptions. Each problem includes a function signature, docstring description, reference implementation, and multiple unit tests, with an average of 7.7 tests per problem. The programming tasks assess comprehension of natural language, reasoning, algorithms, and basic mathematics, at a difficulty level comparable to simple software interview questions. Pass rates are evaluated with the pass@k metric, where k samples are generated per problem and a problem is considered solved if any sample passes all tests. We use all 164 problems for our experiments and a maximum depth limit of 8. For the three questions without sample test cases, we write our own. For value function hyperparameters, we use $\lambda = 0.8$ for the LM score and self-consistency score. For GPT-3.5 we use six internal tests, while for GPT-4 we use four internal tests.

The Mostly Basic Programming Problems (MBPP) \cite{austin2021program} benchmark contains 974 short Python functions designed to evaluate program synthesis techniques. The dataset was constructed by crowdsourcing from workers with basic Python knowledge. Each data point consists of a natural language description of a programming task, a reference solution implementation, and three test cases for functional correctness. The natural language prompts are typically short, one-sentence descriptions. Solutions cover common programming constructs including mathematical operations, list processing, string manipulation, and usage of the Python standard library. On average, solutions are 6.8 lines of code. The dataset is also supplemented with an additional set of 426 problems that were manually verified for unambiguous specifications, standard function signatures, and accurate test cases. We use a randomly selected subset of 397 problems for our experiments. For value function hyperparameters, we use $\lambda = 0.8$ for the LM score and self-consistency score.

\subsection{WebShop}

WebShop \citep{yao2022webshop} is an interactive web-based environment designed to evaluate agents on grounded language understanding and decision-making. It simulates an e-commerce shopping task by providing agents with over 1 million real-world products scraped from Amazon, spanning 5 categories and 113 subcategories. These products contain rich linguistic information, with an average text length of 262 words and a vocabulary size of 224k. In addition, there are over 800k unique product options available for customization. The environment renders webpages in two modes: HTML mode provides pixel-level observations with interactive elements, while simple mode converts the raw HTML into a structured text observation more amenable for training agents. The action space consists of query searches and button clicks, which transition between 4-page types: search, results, item, and item detail. Instructions are crowdsourced natural language specifying product attributes and options, with a total of 12k collected. Automatic rewards are computed by comparing the product purchased by the agent against the attributes and options specified in the instruction, using both lexical matching and semantic similarity metrics. 
\\

\begin{table}[ht]
    \centering
\begin{tabular}{lll}
\toprule
     Type &  Argument &  State $\rightarrow$ Next State \\
\midrule
    \texttt{search} & [\textit{Query}] & Search $\rightarrow$ Results\\ 
    \texttt{choose} & Back to search & $*$ $\rightarrow$ Search\\
    \texttt{choose} & Prev/Next page & Results $\rightarrow$ Results\\
    \texttt{choose} & [\textit{Product title}] & Results $\rightarrow$ Item\\
    \texttt{choose} & [\textit{Option}] & Item $\rightarrow$ Item\\
    \texttt{choose} & Desc/Overview & Item $\rightarrow$ Item-Detail\\
    \texttt{choose} & Previous & Item-Detail $\rightarrow$ Item\\
    \texttt{choose} & Buy & Item $\rightarrow$ Episode End\\
\bottomrule
\end{tabular}
\vspace{5pt}
\caption{Action space of WebShop.}
\label{table:action_space_}
\end{table}

There are two evaluation metrics used in WebShop: (1) \textbf{Task Score} defined as $(100 \times \text{avg. reward})$, which captures the average reward obtained across episodes; and (2) \textbf{Success Rate (SR)} defined as the portion of instructions where $r=1$. The reward is calculated based on the number of attributes satisfied by the selected item. We use 50 environments for our experiments and a maximum depth limit of 15. For value function hyperparameters, we use $\lambda = 0.8$ for the LM score and self-consistency score.

\subsection{Game of 24}

Game of 24 is a mathematical reasoning challenge where the goal is to use basic arithmetic operations to construct 24 out of 4 numbers. We follow the setup from \citet{yao2023tree}, where we measure success if the agent produces a correct equation that equals 24 and uses each input number only once. We report success rate over 50 games. We use $k = 30$ iterations and a maximum depth limit of 5. For value function hyperparameters, we use $\lambda = 0.5$ for the LM score and self-consistency score. Tab.~\ref{table:game24-abl} shows the performance comparison between $\lambda=0.5$ and $\lambda=1$, which validates our design of self-consistency term.

\begin{table}[t]
  \centering
  \begin{tabular}{l|c}
  \toprule
    Prompt Method & Game of 24 (Success Rate) $\uparrow$ \\
    \midrule
    LATS (CoT, $\lambda=1$) & 0.40  \\
    LATS (CoT) & \textbf{0.44}  \\
    \bottomrule
    \end{tabular}
    \caption{Ablations on $\lambda$ in Game of 24 with GPT-3.5. $\lambda = 0.5$ used in the main paper outperforms $\lambda=1$, equivalent to removing self-consistency, which indicates that the self-consistency score improves the performance of LATS.
    }
    \label{table:game24-abl}
    \vspace{-0.1in}
\end{table}

\section{HotPotQA Prompts}
\label{sec:hotpotqa_prompt}
\subsection{Base Acting Prompt}

Solve a question answering task with interleaving Thought, Action, Observation steps. Thought can reason about the current situation, and Action can be three types: 

(1) Search[entity], which searches the exact entity on Wikipedia and returns the first paragraph if it exists. If not, it will return some similar entities to search.

(2) Lookup[keyword], which returns the next sentence containing keyword in the current passage.

(3) Finish[answer], which returns the answer and finishes the task.

After each observation, provide the next Thought and next Action. 

Here are some examples:

Question: Which magazine was started first Arthur's Magazine or First for Women?

Thought 1: I need to search Arthur's Magazine and First for Women, and find which was started first.

Action 1: Search[Arthur's Magazine]

Observation 1: Arthur's Magazine was an American literary periodical published in Philadelphia in the 19th century. Edited by Timothy Shay Arthur, it featured work by Edgar A. Poe, J.H. Ingraham, Sarah Josepha Hale, Thomas G. Spear, and others.[1][2] In May 1846 it was merged into Godey's Lady's Book.[3]

Thought 2: Arthur's Magazine was started in 1844. I need to search First for Women next.

Action 2: Search[First for Women]

Observation 2: First for Women is a woman's magazine published by Bauer Media Group in the USA.[1] The magazine was started in 1989. It is based in Englewood Cliffs, New Jersey.[2] In 2011 the circulation of the magazine was 1,310,696 copies.[3]

Thought 3: First for Women was started in 1989. 1844 (Arthur's Magazine) $<$ 1989 (First for Women), so Arthur's Magazine was started first. 

Action 3: Finish[Arthur's Magazine]

(examples)

You have attempted to answer the following question before and failed. The following reflection(s) give a plan to avoid failing to answer the question in the same way you did previously. Use them to improve your strategy of correctly answering the given question.

(trajectories)

(input)

\subsection{Base Reasoning Prompt}

Solve a question answering task by having a Thought, then Finish with your answer. Thought can reason about the current situation. Finish[answer] returns the answer and finishes the task. You will be given context that you should use to help you answer the question. Start your response with either Action or an indexed Thought

Here are some examples:

Question: What is the elevation range for the area that the eastern sector of the Colorado orogeny extends into? 

Let's think step by step.

Thought 1: The eastern sector of Colorado orogeny extends into the High Plains. 

Thought 2: High Plains rise in elevation from around 1,800 to 7,000 ft

Thought 3: The answer is 1,800 to 7,000 ft.

Action: Finish[1,800 to 7,000 ft]

(examples)

Previous trial:
(trajectories)

(input)

\subsection{Value Function Prompt}

Analyze the trajectories of a solution to a question answering task. The trajectories are labeled by environmental Observations about the situation, Thoughts that can reason about the current situation, and Actions that can be three types: 

(1) Search[entity], which searches the exact entity on Wikipedia and returns the first paragraph if it exists. If not, it will return some similar entities to search.

(2) Lookup[keyword], which returns the next sentence containing keyword in the current passage.

(3) Finish[answer], which returns the answer and finishes the task.

Given a question and a trajectory, evaluate its correctness and provide your reasoning and analysis in detail. Focus on the latest thought, action, and observation. Incomplete trajectories can be correct if the thoughts and actions so far are correct, even if the answer is not found yet. Do not generate additional thoughts or actions. Then at the last line conclude ``Thus the correctness score is {s}'', where s is an integer from 1 to 10.

Question: Which magazine was started first Arthur's Magazine or First for Women?

Thought 1: I need to search Arthur's Magazine and First for Women, and find which was started first.

Action 1: Search[Arthur's Magazine]

Observation 1: Arthur's Magazine was an American literary periodical published in Philadelphia in the 19th century. Edited by Timothy Shay Arthur, it featured work by Edgar A. Poe, J.H. Ingraham, Sarah Josepha Hale, Thomas G. Spear, and others.[1][2] In May 1846 it was merged into Godey's Lady's Book.[3]

This trajectory is correct as it is reasonable to search for the first magazine provided in the question. It is also better to have simple searches corresponding to a single entity, making this the best action.

Thus the correctness score is 10

(other examples)

(failed trajectories)

(context)

\subsection{Reflection Prompt}

Analyze the trajectories of a solution to a question-answering task. The trajectories are labeled by environmental Observations about the situation, Thoughts that can reason about the current situation, and Actions that can be three types: 

(1) Search[entity], which searches the exact entity on Wikipedia and returns the first paragraph if it exists. If not, it will return some similar entities to search.

(2) Lookup[keyword], which returns the next sentence containing keyword in the current passage.

(3) Finish[answer], which returns the answer and finishes the task.

Given a question and a trajectory, evaluate its correctness and provide your reasoning and analysis in detail. Focus on the latest thought, action, and observation. Incomplete trajectories can be correct if the thoughts and actions so far are correct, even if the answer is not found yet. Do not generate additional thoughts or actions. Then at the last line conclude ``Thus the correctness score is {s}'', where s is an integer from 1 to 10.

Question: Which magazine was started first Arthur's Magazine or First for Women?

Thought 1: I need to search Arthur's Magazine and First for Women, and find which was started first.

Action 1: Search[Arthur's Magazine]

Observation 1: Arthur's Magazine was an American literary periodical published in Philadelphia in the 19th century. Edited by Timothy Shay Arthur, it featured work by Edgar A. Poe, J.H. Ingraham, Sarah Josepha Hale, Thomas G. Spear, and others.[1][2] In May 1846 it was merged into Godey's Lady's Book.[3]

This trajectory is correct as it is reasonable to search for the first magazine provided in the question. It is also better to have simple searches corresponding to a single entity, making this the best action.

Thus the correctness score is 10

(other examples)

(failed trajectories)

(context)

\section{Programming Prompts}
\label{sec:programming_prompt}
\subsection{HumanEval function implementation example}
Sample function signature:\\
\begin{lstlisting}
def minSubArraySum(nums):
    Given an array of integers nums, 
    find the minimum sum of any
    non-empty sub-array of nums.
    Example
    minSubArraySum([-1, -2, -3]) == -6
\end{lstlisting}

Sample function body implementation:\\
\begin{lstlisting}
    min_sum = float('inf')
    for i in range(len(nums)):
        current_sum = 0
        for j in range(i, len(nums)):
            current_sum += nums[j]
            if current_sum < min_sum:
                min_sum = current_sum
    return min_sum
\end{lstlisting}

\subsection{Base Acting/Reasoning Prompt}

You are an AI Python assistant. You will be given your previous implementation of a function, a series of unit tests results, and your self-reflection on your previous implementation. Write your full implementation (restate the function signature).

Example 1:

[previous impl]:

\begin{lstlisting}
def add(a: int, b: int) -> int:
    ``Given integers a and b, 
    return the total value of a and b.''
    return a - b
\end{lstlisting}

[unit test results from previous impl]:

Tested passed:

Tests failed:

assert add(1, 2) == 3 \# output: -1

assert add(1, 2) == 4 \# output: -1

[reflection on previous impl]:

The implementation failed the test cases where the input integers are 1 and 2. The issue arises because the code does not add the two integers together, but instead subtracts the second integer from the first. To fix this issue, we should change the operator from `-' to `+' in the return statement. This will ensure that the function returns the correct output for the given input.

[improved impl]:

\begin{lstlisting}
def add(a: int, b: int) -> int:
    ``
    Given integers a and b, 
    return the total value of a and b.
    ''
    return a + b
\end{lstlisting}

\subsection{Reflection Prompt}

You are a Python programming assistant. You will be given a function implementation and a series of unit test results. Your goal is to write a few sentences to explain why your implementation is wrong, as indicated by the tests. You will need this as guidance when you try again later. Only provide the few sentence description in your answer, not the implementation. You will be given a few examples by the user.

Example 1:

[previous impl]:

\begin{lstlisting}
    def add(a: int, b: int) -> int:
        ``
        Given integers a and b, 
        return the total value of a and b.
        ''
    return a - b
\end{lstlisting}

[unit test results from previous impl]:

Tested passed:

Tests failed:

assert add(1, 2) == 3 \# output: -1

assert add(1, 2) == 4 \# output: -1

[reflection on previous impl]:

The implementation failed the test cases where the input integers are 1 and 2. The issue arises because the code does not add the two integers together, but instead subtracts the second integer from the first. To fix this issue, we should change the operator from `-' to `+' in the return statement. This will ensure that the function returns the correct output for the given input.

\subsection{Test Case Generation Prompt}

You are an AI coding assistant that can write unique, diverse, and intuitive unit tests for functions given the signature and docstring.

(input)

\section{WebShop Prompts}
\label{sec:webshop_prompt}
\subsection{Acting Prompt}

Webshop 

Instruction:  

i would like a 3 ounce bottle of bright citrus deodorant for sensitive skin, and price lower than 50.00 dollars 

[Search]  
\\

Action: search[3 ounce bright citrus deodorant sensitive skin]

Observation: 

[Back to Search] 

Page 1 (Total results: 50) 

[Next
] 
[B078GWRC1J] 

Bright Citrus Deodorant by Earth Mama | Natural and Safe for Sensitive Skin, Pregnancy and Breastfeeding, Contains Organic Calendula 3-Ounce 

\$10.99 

[B078GTKVXY] 

Ginger Fresh Deodorant by Earth Mama | Natural and Safe for Sensitive Skin, Pregnancy and Breastfeeding, Contains Organic Calendula 3-Ounce 

\$10.99 

[B08KBVJ4XN] 

Barrel and Oak - Aluminum-Free Deodorant, Deodorant for Men, Essential Oil-Based Scent, 24-Hour Odor Protection, Cedar \& Patchouli Blend, Gentle on Sensitive Skin (Mountain Sage, 2.7 oz, 2-Pack) 

\$15.95  
\\

Action: think[B078GWRC1J and B078GTKVXY are bright citrus deodorant less then 50 dollars. I can check B078GWRC1J first.]

Observation: OK.
\\

Action: click[B078GWRC1J]

Observation: 

[Back to Search] 

[Prev] 

scent [assorted scents][bright citrus][calming lavender][ginger fresh][simply non-scents]
size [travel set (4-pack)][3 ounce (pack of 1)][3-ounce (2-pack)]
Bright Citrus Deodorant by Earth Mama | Natural and Safe for Sensitive Skin, Pregnancy and Breastfeeding, Contains Organic Calendula 3-Ounce 

Price: \$10.99 

Rating: N.A. 

[Description] 

[Features] 

[Reviews] 

[Buy Now]  
\\

Action: think[For 3 ounce bottle of bright citrus deodorant for sensitive skin, the item has options `bright citrus' and `3 ounce (pack of 1)' and seems good to buy.]

Observation: OK.
\\

Action: click[bright citrus]

Observation: You have clicked bright citrus. 
\\

Action: click[3 ounce (pack of 1)]

Observation: You have clicked 3 ounce (pack of 1). 
\\

Action: click[Buy Now]

\subsection{Value Function Prompt}

Given an item to purchase and a trajectory that aims to buy an item that exactly matches the specification, which corresponds to the ideal score of 1.0, analyze the following trajectory, then at the last line conclude, ``Thus the correctness score is {s}'', where s is an integer from 1 to 10.

Here are some examples
\\

Webshop

Instruction:

i am looking for dairy free and apple variety pack of chips, and price lower than 30.00 dollars

[Search]
\\

Action: search[dairy free and apple variety pack of chips]

Observation:

[Back to Search]

Page 1 (Total results: 50)

[Next]

[B07HRFJWP8]

Enjoy Life Foods Soft Baked Ovals, Breakfast Bars, Nut Free Bars, Soy Free, Dairy Free, Non GMO, Gluten Free, Vegan, Variety Pack, 4 Boxes (20 Bars Total)

\$100.0

[B01KMHY5PG]

Enjoy Life Soft Baked Chewy Bars, Variety Pack, Nut Free Bars, Soy Free, Dairy Free, Gluten Free, 6 Boxes (30 Total Bars)

\$21.49

[B008D2X8C4]

Enjoy Life Lentil Chips Variety Pack, Dairy Free Chips, Soy Free, Nut Free, Non GMO, Vegan, Gluten Free, 24 Bags (0.8 oz)

\$100.0
\\

Action: think[The first two items are not what I want. The third item is the dairy free and apple variety pack of chips I want.]

Observation: OK.
\\

Action: click[B008D2X8C4]

Observation:

[Back to Search]

[Prev]

flavor name [dill and sour cream][garlic \& parmesan][light sea salt][margherita pizza][thai chili lime][variety pack]
size [0.8 ounce (pack of 24)][4 ounce (pack of 12)]

Price: \$100.0

Rating: N.A.

[Description]

[Features]

[Reviews]

[Buy Now]
\\

Action: think[The item has options `variety pack' and `0.8 ounce (pack of 24)' and seems good to buy.]

Observation: OK.
\\

Action: click[variety pack]

Observation: You have clicked variety pack.
\\

Action: click[0.8 ounce (pack of 24)]

Observation: You have clicked 0.8 ounce (pack of 24).
\\

Action: click[Buy Now]
\\

Reflection: In this attempt, I was unsuccessful. I accidentally bought a product that was \$100, which is more than my budget of \$30. Either way, the initial search results were not good. Next time, I will do search[``variety pack of chip'';] and then check if the results meet the dairy free and the \$30 budget constraints. I will continue to refine my searches so that I can find more products.

Thus the correctness score is 5
\\

(input)

\subsection{Reflection Prompt}

You are an advanced reasoning agent that can improve based on self-reflection. You will be given a previous reasoning trial in which you were given access to a shopping website and a specific type of item to buy. You were given access to relevant context and an item to purchase. You were unsuccessful in buying the correct item either because you did not find an item meeting all of the required specifications or because you did not select the correct item. The ideal score is 1.0, and anything less is incorrect. In a few sentences, Diagnose a possible reason for failure and devise a new, concise, high-level plan that aims to mitigate the same failure. Use complete sentences.  
Here are some examples:

Previous Trial
Instruction:
i am looking for dairy free and apple variety pack of chips, and price lower than 30.00 dollars
[Search]

Action: search[dairy free and apple variety pack of chips]
Observation:
[Back to Search]
Page 1 (Total results: 50)
[Next $>$]
[B07HRFJWP8]
Enjoy Life Foods Soft Baked Ovals, Breakfast Bars, Nut Free Bars, Soy Free, Dairy Free, Non GMO, Gluten Free, Vegan, Variety Pack, 4 Boxes (20 Bars Total)
\$100.0
[B01KMHY5PG]
Enjoy Life Soft Baked Chewy Bars, Variety Pack, Nut Free Bars, Soy Free, Dairy Free, Gluten Free, 6 Boxes (30 Total Bars)
\$21.49
[B008D2X8C4]
Enjoy Life Lentil Chips Variety Pack, Dairy Free Chips, Soy Free, Nut Free, Non GMO, Vegan, Gluten Free, 24 Bags (0.8 oz)
\$100.0

Action: think[The first two items are not what I want. The third item is the dairy free and apple variety pack of chips I want.]
Observation: OK.

Action: click[B008D2X8C4]
Observation:
[Back to Search]
[$<$ Prev]
flavor name [dill and sour cream][garlic \& parmesan][light sea salt][margherita pizza][thai chili lime][variety pack]
size [0.8 ounce (pack of 24)][4 ounce (pack of 12)]
Price: \$100.0
Rating: N.A.
[Description]
[Features]
[Reviews]
[Buy Now]

Action: think[The item has options `variety pack' and `0.8 ounce (pack of 24)' and seems good to buy.]
Observation: OK.

Action: click[variety pack]
Observation: You have clicked variety pack.

Action: click[0.8 ounce (pack of 24)]
Observation: You have clicked 0.8 ounce (pack of 24).

Action: click[Buy Now]

STATUS: FAIL

Reflection: In this attempt, I was unsuccessful. I accidentally bought a product that was \$100, which is more than my budget of \$30. Either way, the initial search results were not good. Next time, I will do search[``variety pack of chips''] and then check if the results meet the dairy free and the \$30 budget constraints. I will continue to refine my searches so that I can find more products.

Previous Trial
Instruction:
i need gluten free vegetarian smoked peppered bacon - 4 ounce (pack of 2), and price lower than 40.00 dollars
[Search]

Action: search[gluten free vegetarian smoked peppered bacon 4 ounce pack of 2]
Observation:
[Back to Search]
Page 1 (Total results: 50)
[Next $>$]
[B06Y96MXJV]
Smoked Bacon Sea Salt 3-Pack: Smoked Bacon Chipotle, Smoked Bacon and Onion, Smoked Peppered Bacon - All-Natural Sea Salts Slowly Smoked Over Alderwood - No Gluten, No MSG, Non-GMO (12 total oz.)
\$29.99
[B06Y96N1KG]
Spicy Hot Pepper Sea Salt 3-Pack: Ghost Pepper, Jalapeno, Habanero - All Natural, Delicious Blends of Peppers \& Sea Salt - Gluten-Free, Kosher, No MSG, Non-GMO (12 total oz.)
\$29.99
[B07GJTKYJQ]
Louisville Vegan Jerky - 5 Flavor Variety Pack, Non-GMO Soy Protein, Gluten-Free (Black Pepper, Buffalo Dill, Pepperoni, Maple Bacon, and Carolina BBQ, 3 Ounces)
\$42.99

Action: think[The first two items are not what I want. The third item is not what I want, but it is the closest one.]
Observation: OK.

Action: click[B07GJTKYJQ]
Observation:
[Back to Search]
[$<$ Prev]
Louisville Vegan Jerky - 5 Flavor Variety Pack, Non-GMO Soy Protein, Gluten-Free (Black Pepper, Buffalo Dill, Pepperoni, Maple Bacon, and Carolina BBQ, 3 Ounces)
Price: \$42.99
Rating: N.A.
[Description]
[Features]
[Reviews]
[Buy Now]

Action: think[The price is too high and it is not what I want.]
Observation: OK.

Action: click[$<$ Prev]
Observation:
[Back to Search]
Page 1 (Total results: 50)
[Next $>$]
[B06Y96MXJV]
Smoked Bacon Sea Salt 3-Pack: Smoked Bacon Chipotle, Smoked Bacon and Onion, Smoked Peppered Bacon - All-Natural Sea Salts Slowly Smoked Over Alderwood - No Gluten, No MSG, Non-GMO (12 total oz.)
29.99
[B06Y96N1KG]
Spicy Hot Pepper Sea Salt 3-Pack: Ghost Pepper, Jalapeno, Habanero - All Natural, Delicious Blends of Peppers \& Sea Salt - Gluten-Free, Kosher, No MSG, Non-GMO (12 total oz.)
29.99
[B07GJTKYJQ]
Louisville Vegan Jerky - 5 Flavor Variety Pack, Non-GMO Soy Protein, Gluten-Free (Black Pepper, Buffalo Dill, Pepperoni, Maple Bacon, and Carolina BBQ, 3 Ounces)
42.99

Action: think[The first two items are not what I want.]
Observation: OK.

Action: click[Next $>$]
Observation: Invalid action!

Action: think[I need to refine my search.]
Observation: OK.

Action: click[$<$ Back to Search]
Observation: Invalid action!

Action: click[Back to Search]

STATUS: FAIL

Reflection: In this attempt, I was unsuccessful. The initial search results were not the products that I was looking for. Next time, I will do search[``vegetarian bacon''] for a broader search. I will check that the new results will fulfill the gluten free and 4 ounce pack of 2 constraints. I will continue to refine my searches so that I can find more products.

Previous trial:
{trajectory}
Reflection:'''

\end{document}